\documentclass[11pt,a4paper]{article}
\usepackage[hyperref]{AACL-IJCNLP2020}
\usepackage{times}
\usepackage{latexsym}

\usepackage{comment}
\usepackage{url}
\usepackage{amsmath}
\usepackage{amsfonts}
\usepackage{graphicx}
\usepackage{subcaption}
\usepackage{xcolor}
\usepackage{booktabs, multirow} 
\usepackage{soul}
\DeclareMathOperator*{\argmax}{arg\,max}
\DeclareMathOperator*{\argmin}{arg\,min}
\usepackage{footnote}
\makesavenoteenv{tabular}

\usepackage{makecell}

\usepackage{tikz}
\usepackage{tikz-qtree}
\tikzset{edge from parent/.style=
{draw,
edge from parent path={(\tikzparentnode.south)
                    -- +(0, -5pt)
                    -| (\tikzchildnode)}}}
\tikzset{level 1+/.style={level distance=24pt}}
\tikzset{frontier/.style={level distance=20pt}}

\usepackage{todonotes}

\usepackage{microtype}

\aclfinalcopy 

\usepackage{xcolor}
\usepackage{soul}

\newcommand{\redsub}[1]{\small\text{\color{purple} [#1]}}
\newcommand{\subsetting}[1]{$_\text{#1}$}

\title{Heads-up! Unsupervised Constituency Parsing \\
via Self-Attention Heads}

\author{Bowen Li$^\dagger$\quad Taeuk Kim$^\ddagger$\quad Reinald Kim Amplayo$^\dagger$\quad Frank Keller$^\dagger$ \\
$^\dagger$ILCC, School of Informatics, University of Edinburgh, UK \\ 
$^\ddagger$Dept. of Computer Science and Engineering, Seoul National University, Korea \\
\texttt{\{bowen.li,reinald.kim\}@ed.ac.uk} \\
\texttt{taeuk@europa.snu.ac.kr} \quad \texttt{keller@inf.ed.ac.uk} \\
}

\date{}

\begin{document}
\maketitle

\begin{abstract}
  Transformer-based pre-trained language models (PLMs) have
  dramatically improved the state of the art in NLP across many
  tasks. This has led to substantial interest in analyzing the
  syntactic knowledge PLMs learn. Previous approaches to this question
  have been limited, mostly using test suites or probes. Here, we
  propose a novel fully unsupervised parsing approach that extracts
  constituency trees from PLM attention heads. We rank transformer
  attention heads based on their inherent properties, and create an
  ensemble of high-ranking heads to produce the final tree. Our method
  is adaptable to low-resource languages, as it does not rely on
  development sets, which can be expensive to annotate.  Our
  experiments show that the proposed method often outperform existing
  approaches if there is no development set present.
%
%
%
  Our unsupervised parser can also be used as a tool to analyze the
  grammars PLMs learn implicitly.  For this, we use the parse trees
  induced by our method to train a neural PCFG and compare it to a
  grammar derived from a human-annotated treebank.
\end{abstract}

\section{Introduction}

Pre-trained language models (PLMs), particularly BERT
\citep{devlinetal2019bert} and others \citep{yang2019xlnet,
  liu2019roberta, radford2019language} based on the transformer
architecture \citep{vaswani2017attention}, have dramatically improved
the state of the art in NLP.  Such models make it possible to train a
large, generic language model on vast unannotated datasets, and then
fine-tune it for a specific task using a small amount of annotated
data. The success of PLMs has led to a large literature investigating
the linguistic knowledge that PLMs learn implicitly during
pre-training \citep{liuetal2019linguistic, clark2019what,
  kovalevaetal2019revealing, pimentel2020information}, sometimes
referred to as BERTology \citep{rogers2020primer}.

BERTology has been particularly concerned with the question whether
BERT-type models learn syntactic structure.  Typical approaches
include test suites of sentences that instantiate specific syntactic
structures \citep{Goldberg:19}, general probes (also known as
diagnostic classifiers, \citealt{belinkov2019analysis}) or structural
probes \citep{hewitt2019structural}.  All of these approaches are
limited: the first one requires the laborious compilation of language-
and construction-specific suites of sentences; the second one
sometimes fails to adequately reflect differences in representations
\citep{zhangbowman2018language, hewitt-liang-2019-designing,
  voita2020information}; the third one involves designing a novel
extraction model that is not applicable to tasks other than probing
\citep{Maudslay:20}.

It is therefore natural to use a parsing task to test whether the
representations learned by PLMs contain usable syntactic
information. This enables us to test syntactic structure in general,
rather than specific constructions, and doesn't require a specialized
probe. In this paper, we will therefore use PLM attention heads to
construct an \emph{unsupervised constituency parser.}  Previously,
related approaches have been proposed under the heading of
\emph{zero-shot} constituency parsing \citep{Kim2020Are,
  kim2020multilingual}.\footnote{Like \citet{kim2020multilingual}, we
  use \emph{zero-shot} to refer to the transfer from language modeling
  to constituency parsing.}  However, this prior work crucially relies
on an annotated development set in order to identify transformer heads
that are sensitive to syntactic structure. Existing approaches
therefore are not truly unsupervised.  For most low resource
languages, no such annotated data is available, and often not even an
annotation scheme exists.  Thus, assuming a development set is not a
realistic experimental setup \citep{kann2019towards}.
If a suitable development set is available, \citet{shi2020role} shows that an existing supervised parser trained on a few-shot setting can outperform all the unsupervised parsing methods by a significant margin.
It strongly challenges tuning on an annotated development set for unsupervised parsing.

In this paper, we propose a novel approach to build a PLM-based
unsupervised parser that does not require a development set: we rank
transformer heads based on their inherent properties, such as how
likely tokens are to be grouped in a hierarchical structure.  We then
ensemble the top-$K$ heads to produce constituency trees.

We evaluate our approach and previous zero-shot approaches on the
English Penn Treebank (PTB) and eight other languages on the SPMRL
dataset.  On the one hand, if the development set is absent, our
approach largely outperforms previous zero-shot approaches on the
English PTB.  On the other hand, if previous zero-shot approaches are
equipped with the development set, our approach can still match the
parsing performance of these approaches using the single best head or
layer-wise ensembling.  For the multilingual experiment, we take
advantage of the top-$K$ heads selected in English and directly parse
other languages using our approach.  Surprisingly, on five out of nine
languages, this \emph{crosslingual} unsupervised parser matches
previous approaches that rely on a development set in each target
language with the single best head or layer-wise ensembling.  However,
our fully unsupervised method lags behind the previous
state-of-the-art zero-shot parser if a top-$K$ ensemble is used.

Furthermore, our approach can be use as a tool to analyze the
capability of PLMs in learning syntactic knowledge.  As no human
annotation is required, our approach has the potential to
reveal the grammar PLMs learn implicitly.  Here, we use the tree
structures generated by our parser to train a neural PCFG.  We
evaluate the learned grammar against the English PTB on internal tags
and production rules both qualitatively and quantitatively.

\section{Related Work}

Recently, neural models have renewed interest in grammar induction.
Earlier work \citep{choi2018learning, williamsetal2018latent}
attempted to induce grammar by optimizing a sentence classification
objective, while follow-up work \citep{htutetal2018grammarinduction,
  shen2018prpn, shen2019ordered} showed that a language modeling
objective performs better.  Latest work employed autoencoders or
probabilistic grammars \citep{drozdovetal2019unsupervised,
  kimetal2019compound, kimetal2019unsupervised, zhu2020return}.

A new line of work is zero-shot constituency parsing, whose goal is to
automatically extract trees from PLMs in a parameter-free fashion.
The top-down zero-shot parser \citep{Kim2020Are} utilizes the concept
of \emph{syntactic distance} \citep{shen2018straight}, where trees are
induced by an algorithm that recursively splits a sequence of words in
a top-down manner.  However, this approach suffers from its greedy
search mode, failing to take into account all possible subtrees.  The
chart-based zero-shot parser \citep{kim2020multilingual} applies chart
parsing to address this problem.
\citet{wu2020perturbed} introduced a parameter-free probing technique
to analyze PLMs via perturbed masking.

There is also prior work on extracting constituency trees from
self-attention mechanisms of transformers.
\citet{marevcek2018extracting} proposed heuristic approaches to
convert attention weights to trees.
\citet{marecekrosa2019balustrades} introduced a chart-based tree
extraction method in transformer-based neural machine translation encoders and provide a
quantitative study.

\section{Zero-shot Constituency Parsing via PLMs}

In this section, we briefly review the chart-based zero-shot parser and then introduce our ranking-based zero-shot parser.

\subsection{Chart-based Zero-shot Parsing} 
\label{subsec: chart-based zero-shot constituency parsing}

In chart-based zero-shot parsing, a real-valued score 
$s_{tree}(\boldsymbol{t})$ is assigned for each tree candidate $\boldsymbol{t}$, which decomposes as:
\begin{equation*}
    s_{tree}(\boldsymbol{t}) = \sum_{(i,j) \in \boldsymbol{t}} s_{span}(i, j),
\end{equation*}
where $s_{span}(i, j)$ is the score (or cost) for a constituent that is located between positions $i$ and $j$ ($1 \le i \le j \le n$, where $n$ is the length of the sentence).
Specifically, for a span of length 1, $s_{span}(i, j)$ is defined as 0 when $i = j$.
For a span longer than 1, the following recursion applies:
\begin{gather}
  \resizebox{.89\linewidth}{!}{$
    s_{span} (i, j) = s_{comp}(i,j) + \min_{i \le k < j} s_{split}(i, k, j)
  $} \label{eq: recursive span} \\
  \resizebox{.87\linewidth}{!}{$
  s_{split} (i, k, j) = s_{span}(i, k) + s_{span}(k+1, j),
$} \label{eq: recursive split}
\end{gather}
where $s_{comp}(\cdot, \cdot)$ measures the validity or compositionality of the span $(i,j)$ itself, while $s_{split}(i, k, j)$ indicates how plausible it is to split the span $(i,j)$ at position $k$. 
Two alternatives have been developed in \citet{kim2020multilingual} for $s_{comp}(\cdot, \cdot)$: the pair score function $s_{p}(\cdot, \cdot)$ and the characteristic score function $s_{c}(\cdot, \cdot)$. 

The pair score function $s_{p}(\cdot, \cdot)$ computes the average pair-wise distance in a given span:
\begin{equation}
  \label{eq: s_p}
  \resizebox{0.891\linewidth}{!}{$
  \begin{split}
  s_{p} (i,j) =
  \frac{1}{\binom{j-i+1}{2}} \sum_{\substack{ (w_x,w_y) \in \texttt{pair}(i,j) } } f(g(w_x), g(w_y)),
  \end{split}
  $}
\end{equation}
where $\texttt{pair}(i,j)$ returns a set consisting of all combinations of two words (e.g.,~$w_x$, $w_y$) inside the span $(i,j)$.

Functions $f(\cdot, \cdot)$ and $g(\cdot)$ are the distance measure function and the representation extractor function, respectively. 
For $g$, given $l$ as the number of layers in a PLM, $g$ is actually a set of functions $g = \{g_{(u,v)}^d|u=1,\dots,l, v=1,\dots,a\}$, each of which outputs the attention distribution of the $v^{th}$ attention head on the $u^{th}$ layer of the PLM.\footnote{
  The hidden representations of the given words can also serve as an alternative for $g$. 
  But \citet{Kim2020Are} show that the attention distributions provide more syntactic clues under the zero-shot setting.
}
In case of the function $f$, there are also two options, Jensen-Shannon (JSD) and Hellinger (HEL) distance. Thus, $f = \{\text{JSD}, \text{HEL}\}$.

The characteristic score function $s_{c}(\cdot,\cdot)$ measures the distance between each word in the constituent and a predefined characteristic value $\boldsymbol{c}$ (e.g.,~the center of the constituent):
\begin{equation}\label{eq: s_c}
    s_{c}(i,j) = \frac{1}{j-i+1} \sum_{i\le x \le j} f(g(w_x), \boldsymbol{c}),
\end{equation}
where $\boldsymbol{c} = \frac{1}{j-i+1} \sum_{i \le y \le j} g(w_y)$.

Since $s_{comp}(\cdot, \cdot)$ is well defined, it is straightforward to compute every possible case of  $s_{span}(i,j)$ using the CKY algorithm \citep{cocke1969programming, kasami1966efficient, younger1967recognition}.
Finally, the parser outputs $\hat{\boldsymbol{t}}$, the tree that requires the lowest score (cost) to build, as a prediction for the parse tree of the input sentence: $\hat{\boldsymbol{t}} = \argmin_{\boldsymbol{t}}s_{tree}(\boldsymbol{t})$.

For attention heads ensembling, both a layer-wise ensemble and a top-$K$ ensemble are considered.
The first one averages all attention heads from a specific layer, while the second one averages the top-$K$ heads from across different layers.
At test time, separate trees produced by different heads are merged to one final tree via \emph{syntactic distance}.\footnote{
  Details can be found in \citet{kim2020multilingual}.
  For the ensemble parsing, marrying chart-based parser and top-down parser yields better results than averaging the attention distributions.
}
The chart-based zero-shot parser achieves the state of the art in zero-shot constituency parsing.

\subsection{Ranking-based Zero-shot Parsing} 
\label{subsec: ranking-parser}

The chart-based zero-shot parser relies on the existing development set of a treebank (e.g.,~the English PTB) to select the best configuration, i.e., the combination of $\{g\ |\ g_{(u,v)}^d, u=1,\dots,l, v=1,\dots,a\}$, $\{f\ |\ \text{JSD}, \text{HEL}\}$, $\{s_{comp}\ |\ s_{p}, s_{c}\}$, and heads ensemble that achieves the best parsing accuracy.
Such a development set always contains hundreds of sentences, hence considerable annotation effort is still required.
From the perspective of unsupervised parsing, such results arguably are not fully unsupervised.\footnote{
  Some previous work \citep{shen2018prpn, shen2019ordered, drozdovetal2019unsupervised, kimetal2019compound} also use a development set to tune hyperparameters or early-stop training.
} 
Another argument against using a development set is that the linguistic assumptions inherent in the expert annotation required to create the development set potentially restrict our exploration of how PLMs model the constituency structures. 
It could be that the PLM learns valid constituency structures, which however do not match the annotation guidelines that were used to create the development set.

Here, we take a radical departure from the previous work in order to extract constituency trees from PLMs in a fully unsupervised manner. 
We propose a two-step procedure for unsupervised parsing: (1)~identify syntax-related attention heads directly from PLMs without relying on a development set of a treebank; (2)~ensemble the selected top-$K$ heads to produce the constituency trees.

For identification of the syntax-related attention heads, we rank all heads by scoring them with a chart-based ranker.
We borrow the idea of the chart-based zero-shot parser to build our ranker.
Given an input sentence and a specific choice of $f$ and $s_{comp}$, each attention head $g_{(u,v)}^d$ in the PLM yields one unique attention distribution.
Using the chart-based zero-shot parser in Section~\ref{subsec: chart-based zero-shot constituency parsing}, we can obtain the score of the best constituency tree as:\footnote{
  Our ranking method works approximately as a maximum a posteriori probability (MAP)
  estimate, since we only consider the best tree the attention head generates. 
  In unsupervised parsing, marginalization is a standard method for model development.
  We have tried to apply marginalization to our ranking algorithm where all possible trees are considered and the sum score is calculated (using the \texttt{logsumexp} trick) for ranking.
  But marginalization does not work well for attention distributions, where an ``attending broadly'' head with higher entropy is more favorable than a syntax-related head with lower entropy.
  So we only consider the score of the best tree.}
\begin{equation}
\label{eq: score}
\resizebox{0.89\linewidth}{!}{$
  s_{parsing}(u,v) = s_{tree}(\hat{\boldsymbol{t}}) = \sum_{(i,j) \in \hat{\boldsymbol{t}}} s_{span}(i, j),
  $}
\end{equation}
where $\hat{\boldsymbol{t}} = \argmin_{\boldsymbol{t}}s_{tree}(\boldsymbol{t})$. 
It is obvious that all combinations of $\{f\ |\ \text{JSD}, \text{HEL}\}$ and $\{s_{comp}\ |\ s_{p}, s_{c}\}$ will produce multiple scores for a given head. 
Here we average the scores of all such combinations to get one single score. 
Then we rank all attention heads and select the syntax-related heads for parsing.
However, directly applying the chart-based zero-shot parser in Section~\ref{subsec: chart-based zero-shot constituency parsing} for ranking delivers a trivial, ill-posed solution.
The recursion in Eq.~(\ref{eq: recursive split}) only encourages the intra-similarity inside the span. 
Intuitively, one attention head that produces the \emph{same} attention distribution for  each token (e.g.,~a uniform attention distribution or one that forces every token to attend to one specific token) will get the lowest score (cost) and the highest ranking.\footnote{
  Such cases do exist in PLMs. \citet{clark2019what} shows that BERT exhibits clear surface-level attention patterns. 
  Some of these patterns will deliver ill-posed solutions in ranking: attend broadly, attend to a special tokens (e.g.,~\texttt{[SEP]}), attend to punctuation (e.g.,~period). 
  One can also observe these patterns using the visualization tool provided by \citet{vig2019multiscale}.
}

To address this issue, we first introduce inter-similarity into the recursion in Eq.~(\ref{eq: recursive split}) and get the following:
\begin{equation}
  \resizebox{0.89\linewidth}{!}{$
  \begin{split}
    s_{split}&(i, k, j) = & \\
    \ \ \ \ \ \ \ & s_{span}(i, k) + s_{span}(k+1, j) - s_{cross}(i, k, j),
  \end{split}
  $}
\end{equation}
where the cross score $s_{cross}(i, k, j)$ is the similarity between two subspans $(i,k)$ and $(k+1, j)$. 
However, this formulation forces the algorithm to go to the other extreme: one attention head that produces a totally \emph{different} distribution for each token (e.g.,~force each token to attend to itself or the previous/next token) will get the highest ranking.
To balance the inter- and intra-similarity and avoid having to introduce a tunable coefficient, we simply add a length-based weighting term to Eq.~(\ref{eq: recursive span}) and get:
\begin{equation}
  \resizebox{0.89\linewidth}{!}{$
  \begin{split}
    s_{span}&(i, j) = & \\
    \ \ \ \ \ \ \ & \frac{j - i + 1}{n} ( s_{comp}(i,j) + \min_{i \le k < j} s_{split}(i, k, j) ), \\
  \end{split}
  $}
\end{equation}
where $j - i + 1$ is the length of the span $(i, j)$. 
The length ratio functions as a regulator to assign larger weights to longer spans.
This is motivated by the fact that longer constituents should contribute more to the scoring of the parse tree, since the inter-similarity always has strong effects on shorter spans.
In this way, the inter- and intra-similarity can be balanced.

With respect to the choice for $s_{cross}(i, k, j)$, we follow the idea of $s_{p}$ and $s_{c}$ in Eq.~(\ref{eq: s_p}) and (\ref{eq: s_c}) and propose the pair score function $s_{p{\text x}}$ and the characteristic score function $s_{c{\text x}}$\footnote{
 Subscripts in the naming of functions in this paper: $p$ -- pair score, $c$ -- characteristic score, $\text{x}$ -- cross score.
} 
for cross score computation. $s_{p{\text x}}$ is defined as:
\begin{equation*}
\begin{split}
s_{p{\text x}}(i,j) & = \frac{1}{(k - i + 1)(j - k)} \\
&\sum_{(w_x,w_y)\in \texttt{prod}(i,k,j)} f(g(w_x), g(w_y)),
\end{split}
\end{equation*}
where $\texttt{prod}(i,k,j)$ returns a set of the product of words from the two subspans $(i,k)$ and $(k+1, j)$.
And $s_{c{\text x}}$ is defined as:
\begin{equation*}
s_{c{\text x}}(i, j) = f(\boldsymbol{c}_{i, k}, \boldsymbol{c}_{k+1, j}),
\end{equation*}
where $\boldsymbol{c}_{i, k} = \frac{1}{k-i+1} \sum_{i \le x \le k} g(w_x)$, $\boldsymbol{c}_{k + 1, j} = \frac{1}{j-k} \sum_{k + 1 \le y \le j} g(w_y)$.

We average all the combinations of $\{f\ |\ \text{JSD}, \text{HEL}\}$, $\{s_{comp}\ |\ s_{p}, s_{c}\}$ and $\{s_{cross}\ |\ s_{p{\text x}}, s_{c{\text x}}\}$ to rank all the attention heads and select the top-$K$ heads.
After the ranking step, we perform constituency parsing by ensembling the selected heads.
We simply employ the ensemble method in Section~\ref{subsec: chart-based zero-shot constituency parsing} and average all the combinations of $\{f\ |\ \text{JSD}, \text{HEL}\}$ and $\{s_{comp}\ |\ s_{p}, s_{c}\}$ to get a single predicted parse tree for a given sentence.

\subsection{How to select $K$}
\label{subsec: k selection}

For ensemble parsing, \citet{kim2020multilingual} proposed three settings: the best head, layerwise ensemble, and top-$K$ ensemble.
To prevent introducing a tunable hyperparameter, we propose to select a value for $K$ dynamically based on a property of the ranking score in Eq.~(\ref{eq: score}).

Since we use a similarity-based distance, the lower the ranking score, the higher the ranking.
Assuming that scores are computed for all attention heads, we can sort the scores in ascending order.
Intuitively, given the order, we would like to choose the $k$ for which ranking score increases the most, which means syntactic relatedness drops the most.
Suppose $s_{parsing}(k)$ is the ranking score where $k$ is the head index in the ascending order,
then this is equivalent to finding the $k$ with the greatest gradient on the curve of the score.
We first estimate the gradient of $s_{parsing}(k)$ and then find the $k$ with the greatest gradient.
Finally, $K$ is computed as:
\begin{equation*}
  \resizebox{1.0\linewidth}{!}{$
  \begin{split}
  K =& \argmax_k \sum_{\substack{ {k - \delta \le j \le k + \delta} \\ {j \ne k}}} \frac{s_{parsing}(k + j) - s_{parsing}(k)}{j},  	
  \end{split}
  $}
\end{equation*}
where we smooth the gradient by considering $\delta$ steps. Here, we set~$\delta = 3$.

In practice, we find that the greatest gradient always happens in the head or the tail of the curve. 
For the robustness, we select the $K$ from the middle range of the score function curve, i.e., starting from 30 and ending with 75\% of all heads.\footnote{
Although our ranking algorithm can filter out \emph{noisy} heads, by observing the attention heatmaps, we find that noisy heads sometimes still rank high. 
We do not do any post-processing to further filter out the noisy heads, so we empirically search $k$ starting at~30.}
We also provide a \emph{lazy} option for $K$ selection, which simply assume a fixed value of 30 for the top-$K$ ensemble.	

\section{Grammar Learning}

We are also interested in exploring to what extent the syntactic knowledge acquired by PLMs resembles human-annotated constituency grammars. 
For this exploration, we infer a constituency grammar, in the form of probabilistic production rules, from the trees induced from PLMs. This grammar can then be analyzed further, and compared to human-derived grammars.
Thanks to the recent progress in neural parameterization, neural PCFGs have been successfully applied to unsupervised constituency parsing \citep{kimetal2019compound}.
We harness this model\footnote{
  A more advanced version of the neural PCFG, the compound PCFG, has also been developed in \citet{kimetal2019compound}.
  In this model variant, a compound probability distribution is built upon the parameters of a neural PCFG.
  In  preliminary experiments, we found the compound PCFG learns similar grammars as the neural PCFG.
  So we only use the more light-weight neural PCFG in this work.}
to learn probabilistic constituency grammars from PLMs by maximizing the joint likelihood of sentences and parse trees induced from PLMs.
In the following, we first briefly review the neural PCFG and then introduce our training algorithm.

\subsection{Neural PCFGs}

A probabilistic context-free grammar (PCFG) consists of a 5-tuple grammar $\mathcal{G}=(S, \mathcal{N}, \mathcal{P}, \Sigma, \mathcal{R})$ and rule probabilities $\pi=\left\{\pi_{r}\right\}_{r \in \mathcal{R}}$, where $S$ is the start symbol, $\mathcal{N}$ is a finite set of nonterminals, $\mathcal{P}$ is a finite set of preterminals, $\Sigma$ is a finite set of terminal symbols, and $\mathcal{R}$ is a finite set of rules associated with probabilities $\pi$.
The rules are of the form:
\begin{equation*}
  \begin{array}{lll}
    S \rightarrow A, && A \in \mathcal{N} \\
    A \rightarrow B C, && A \in \mathcal{N}, \quad B, C \in \mathcal{N} \cup \mathcal{P} \\
    T \rightarrow w, && T \in \mathcal{P}, w \in \Sigma .
    \end{array}
\end{equation*}
Assuming $\mathcal{T}_{\mathcal{G}}$ is the set of all possible parse trees of $\mathcal{G}$, the probability of a parse tree $\boldsymbol{t} \in \mathcal{T}_{\mathcal{G}}$ is defined as $p(\boldsymbol{t})=\prod_{r \in t_{\mathcal{R}}} \pi_{r}$, where $t_{\mathcal{R}}$ is the set of rules used in the derivation of $\boldsymbol{t}$.
A PCFG also defines the probability of a given sentence $\boldsymbol{x}$ (string of terminals $\boldsymbol{x} \in \Sigma^*$) via $p(\boldsymbol{x})=\sum_{\boldsymbol{t} \in \mathcal{T}_{\mathcal{G}}(\boldsymbol{x})} p(\boldsymbol{t})$, where $\mathcal{T}_{\mathcal{G}}(\boldsymbol{x})=\{\boldsymbol{t} | \texttt{yield} (\boldsymbol{t})=\boldsymbol{x}\}$, i.e., the set
of trees $\boldsymbol{t}$ such that $\boldsymbol{t}$’s leaves are $\boldsymbol{x}$.

The traditional way to parameterize a PCFG is to assign a scalar to each rule $\pi_r$ under the constraint that valid probability distributions must be formed.
For unsupervised parsing, however, this parameterization has been shown to be unable to learn meaningful grammars from natural language data \citep{carroll1992two}.
Distributed representations, the core concept of the modern deep learning, have been introduced to address this issue \citep{kimetal2019compound}.
Specifically, embeddings are associated with symbols and rules are modeled based on such distributed and shared representations.

In the neural PCFG, the log marginal likelihood:
\begin{equation*}
  \log p_{\theta}(\boldsymbol{x})=\log \sum_{\boldsymbol{t} \in \mathcal{T}_{\mathcal{G}}(\boldsymbol{x})} p_{\theta}(\boldsymbol{t})
\end{equation*}
can be computed by summing out the latent parse trees using the inside algorithm \citep{baker1979trainable}, which is differentiable and amenable to gradient based optimization.
We refer readers to the original paper of \citet{kimetal2019compound} for details on the model architecture and training scheme.

\subsection{Learning Grammars from Induced Trees} \label{subsec: learn grammar}

Given the trees induced from PLMs (described in Section~\ref{subsec: ranking-parser}), we use neural PCFGs to learn constituency grammars.
In contrast to unsupervised parsing, where neural PCFGs are trained solely on raw natural language data, we train them on the sentences and the corresponding tree structures induced from PLMs.
Note that this differs from a fully supervised parsing setting, where both tree structures and internal constituency tags (nonterminals and preterminals) are provided in the treebank.
In our case, the trees induced from PLMs have no internal annotations.

For the neural PCFG training, the joint likelihood is given by:
\begin{equation*}
  \log p(\boldsymbol{x}, \hat{\boldsymbol{t}})=\sum_{r \in \hat{t}_{\mathcal{R}}} \log \pi_{r},
\end{equation*}
where $\hat{\boldsymbol{t}}$ is the induced tree and $\hat{t}_{\mathcal{R}}$ is the set of rules applied in the derivation of $\hat{\boldsymbol{t}}$. 
Although tree structures are given during training, marginalization is still involved: all internal tags will be marginalized to compute the joint likelihood.
Therefore, the grammars learned by our method are anonymized: nonterminals and preterminals will be annotated as NT-$id$ and T-$id$, respectively, where $id$ is an arbitrary ID number.

\section{Experiments}

We conduct experiments to evaluate the unsupervised parsing performance of our ranking-based zero-shot parser on English and eight other languages (Basque, French, German, Hebrew, Hungarian, Korean, Polish, Swedish).
For the grammars learned from the induced parse trees, we perform qualitative and quantitative analysis on how the learned grammars resemble the human-crafted grammar of the English PTB. 

\subsection{General Setup}

\begin{figure*}[!htp]
  \centering
  \begin{subfigure}[b]{0.31\textwidth}
      \centering
      \includegraphics[width=\textwidth]{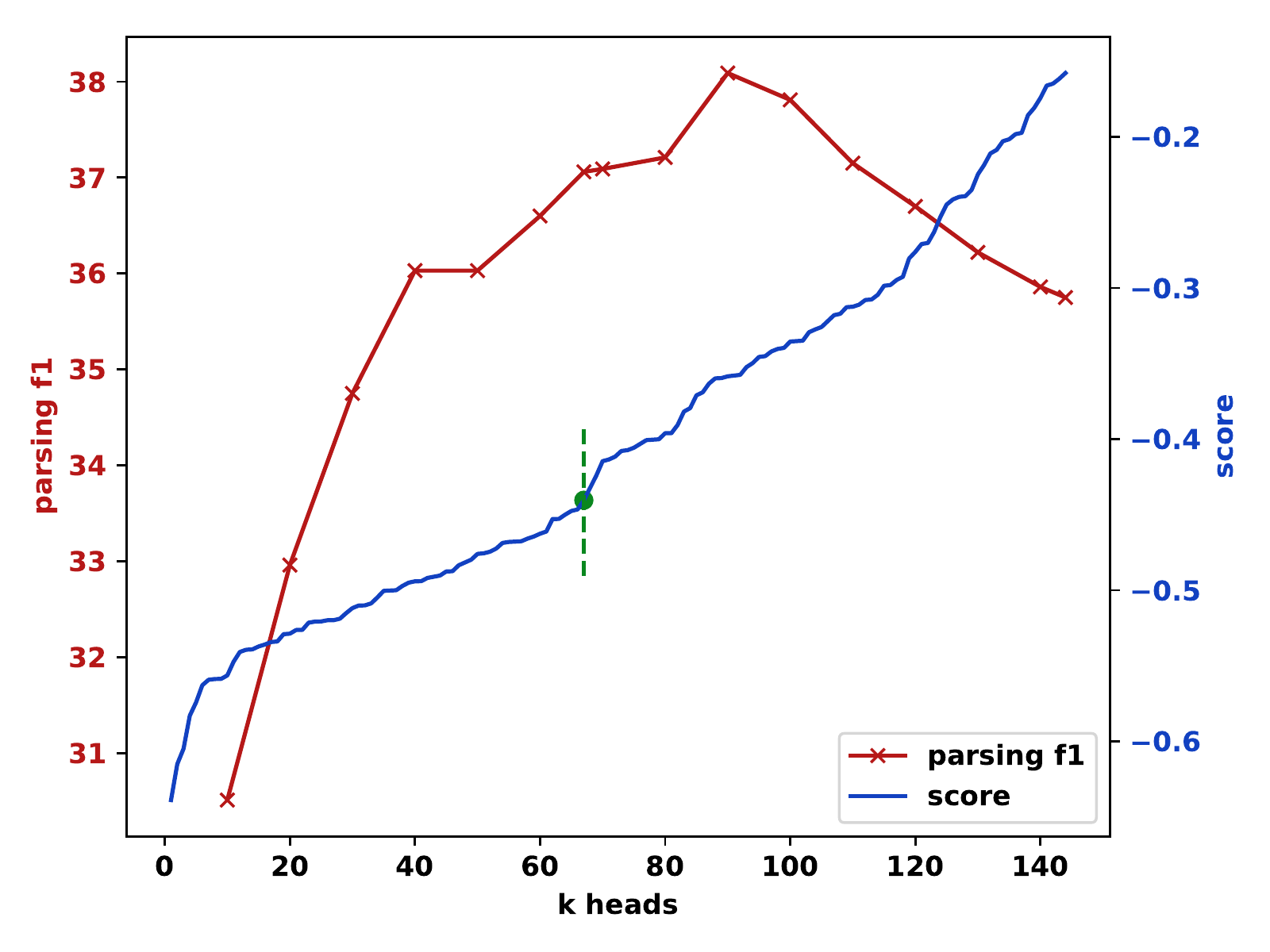}
      \caption[Network2]%
      {{\small BERT-base-cased}}    
      \label{fig:bert-base-cased}
  \end{subfigure} \hfil
  \begin{subfigure}[b]{0.31\textwidth}  
      \centering 
      \includegraphics[width=\textwidth]{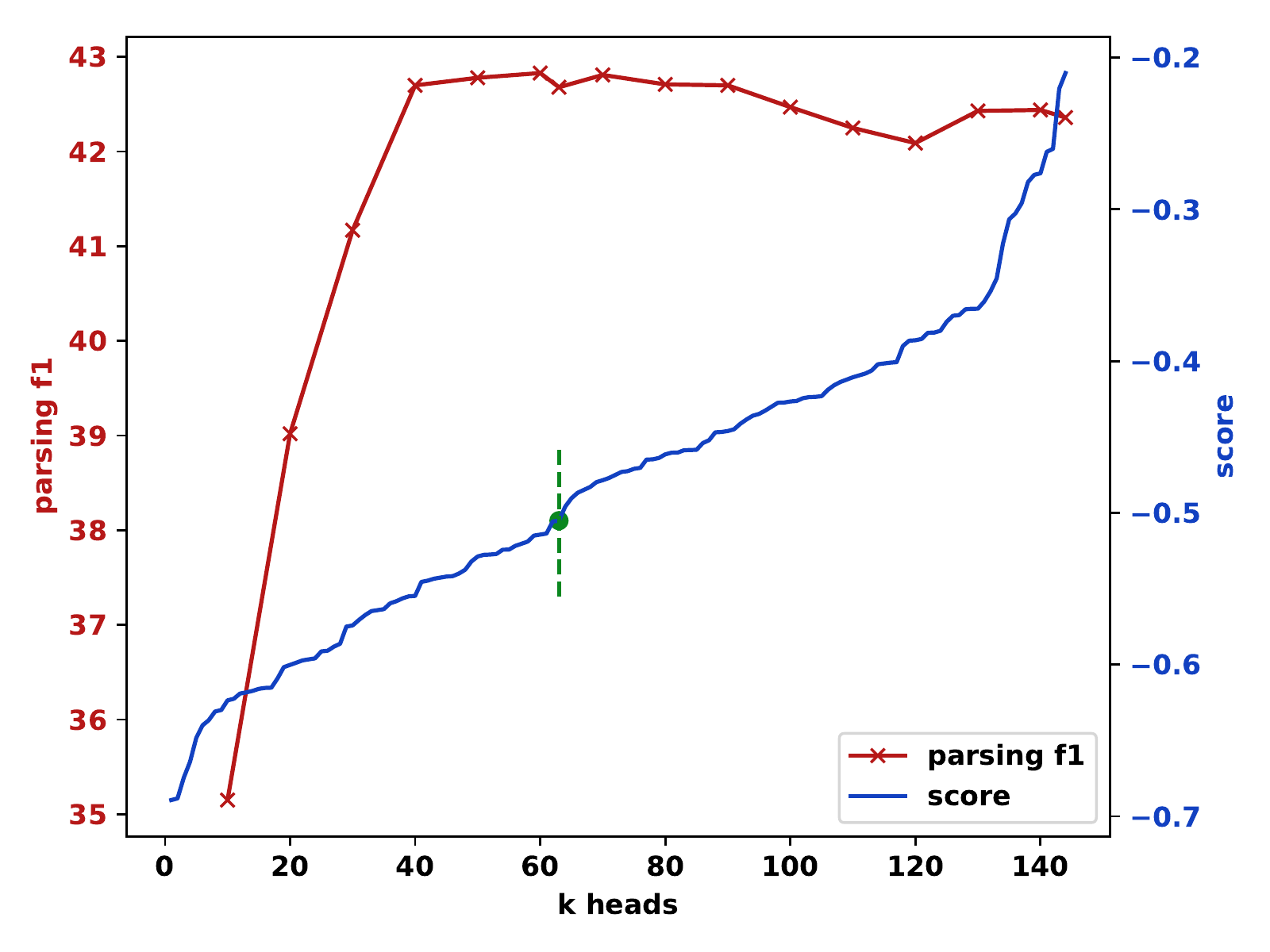}
      \caption[]%
      {{\small XLNet-base-cased}}    
      \label{fig:xlnet-base-cased}
  \end{subfigure} \hfil
  \begin{subfigure}[b]{0.31\textwidth}   
      \centering 
      \includegraphics[width=\textwidth]{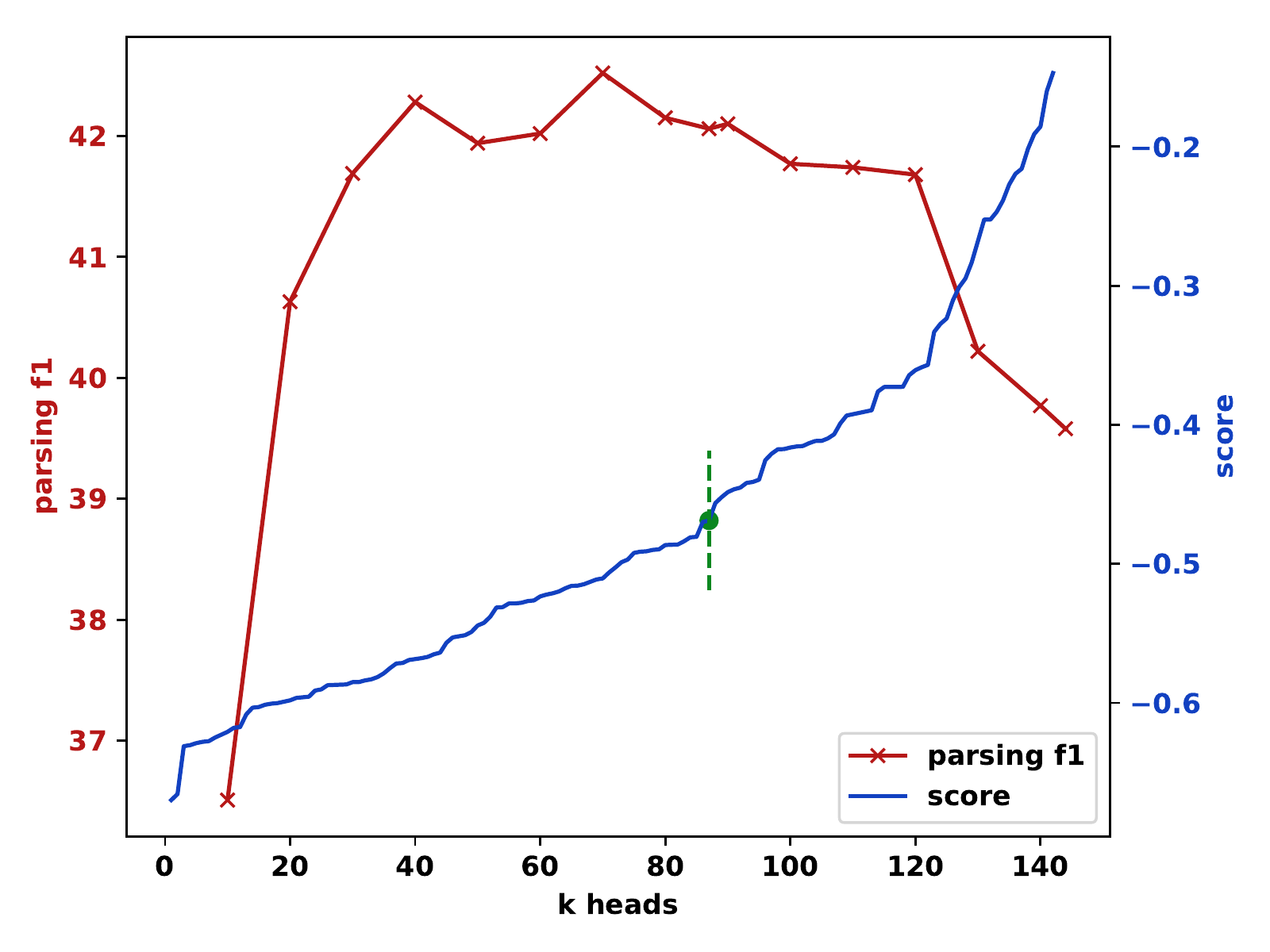}
      \caption[]%
      {{\small RoBERTa-base}}    
      \label{fig:roberta-base}
  \end{subfigure} 

  \begin{subfigure}[b]{0.31\textwidth}   
    \centering 
    \includegraphics[width=\textwidth]{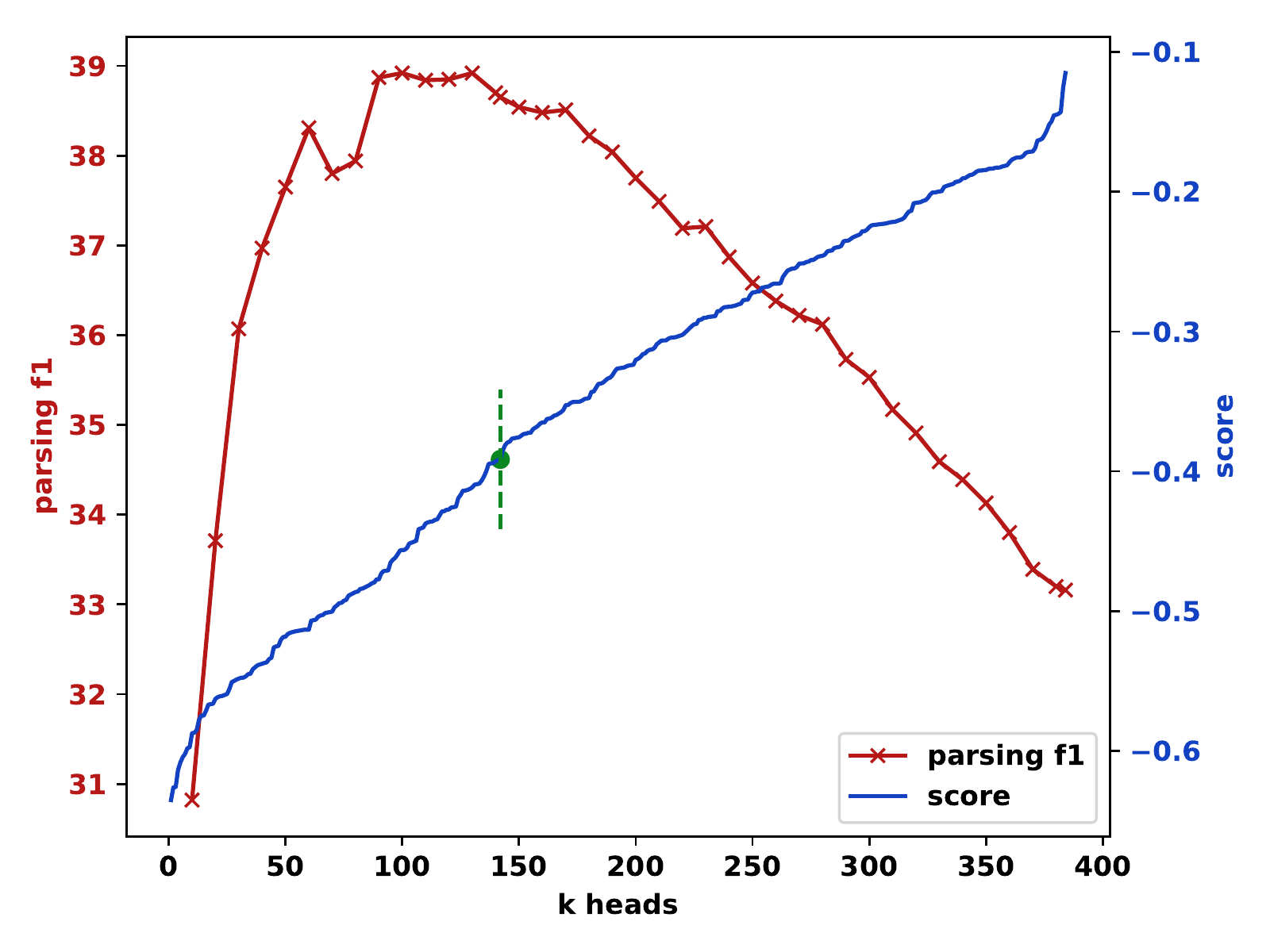}
    \caption[]%
    {{\small BERT-large-cased}}    
    \label{fig:bert-large-cased}
  \end{subfigure} \hfil
  \begin{subfigure}[b]{0.31\textwidth}   
    \centering 
    \includegraphics[width=\textwidth]{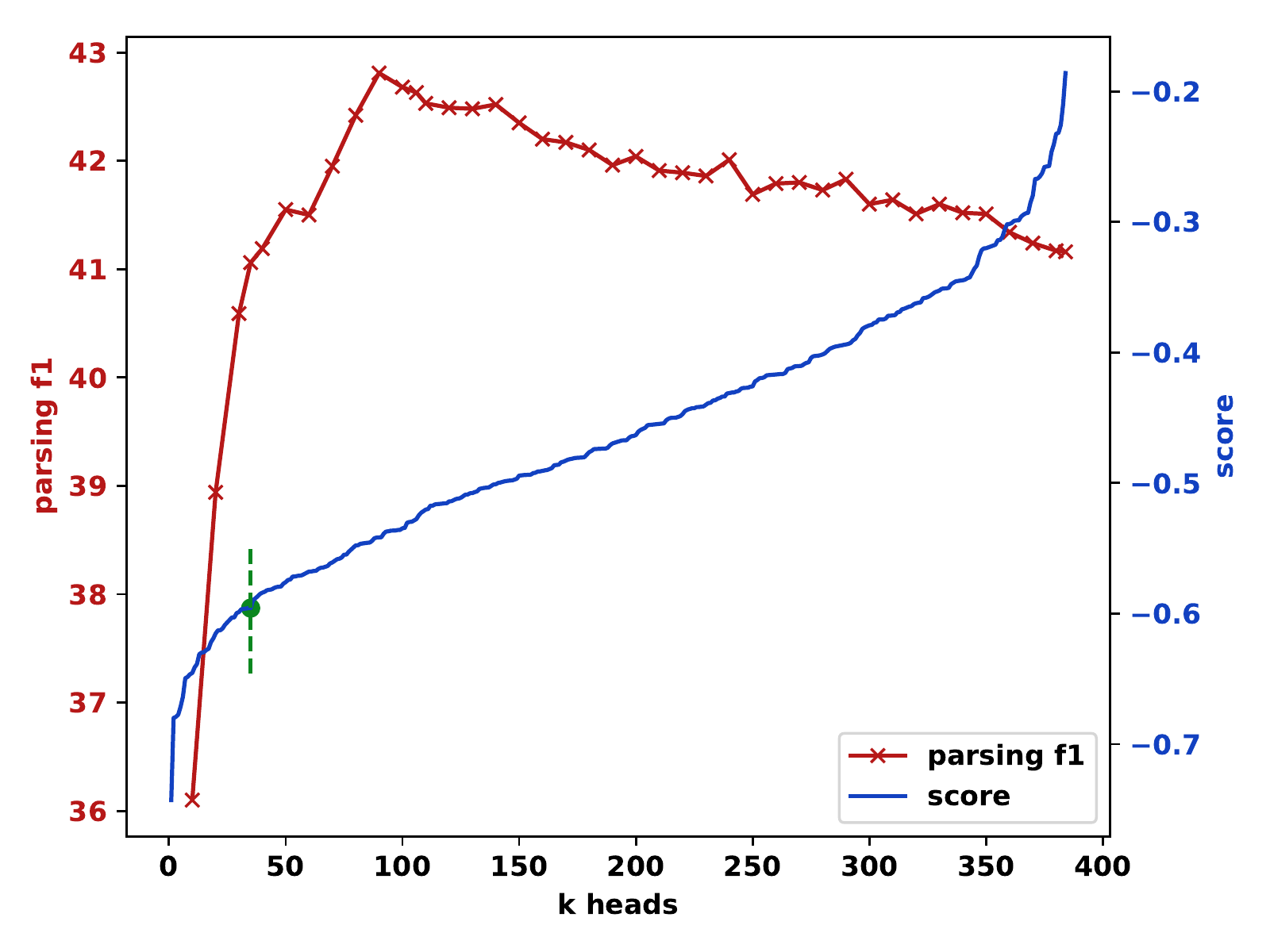}
    \caption[]%
    {{\small XLNet-large-cased}}    
    \label{fig:xlnet-large-cased}
  \end{subfigure} \hfil
  \begin{subfigure}[b]{0.31\textwidth}   
    \centering 
    \includegraphics[width=\textwidth]{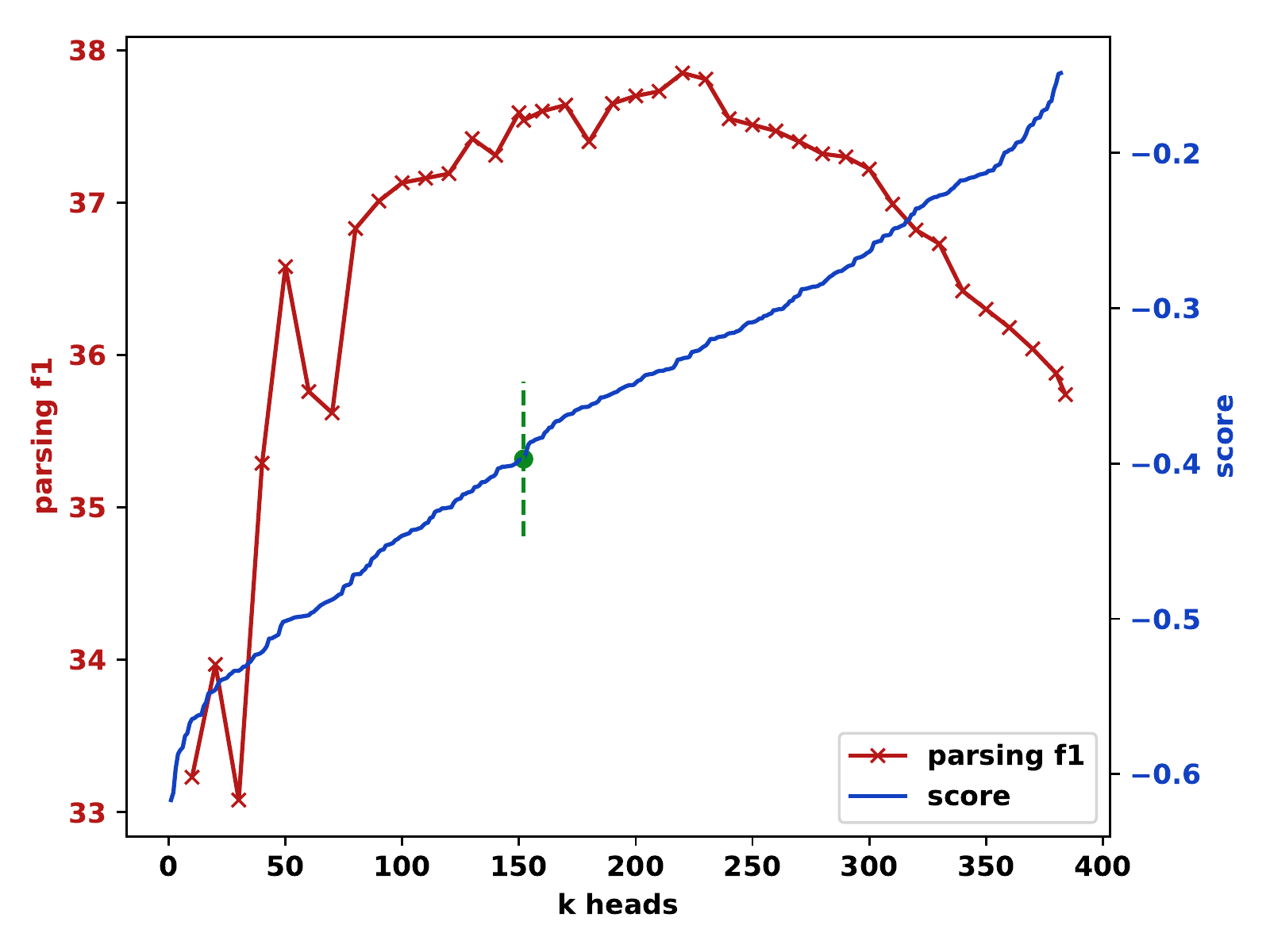}
    \caption[]%
    {{\small RoBERTa-large}}    
    \label{fig:roberta-large}
  \end{subfigure}

  \caption{ 
  Relation between $K$ for top-$K$ and parsing performance on different PLMs. 
  The blue curve shows the ranking score of heads where heads are sorted in an ascending order.
  The red curve shows the parsing performance that is evaluated on the PTB test set given every 10 heads.
  The green dashed line indicates the dynamic~$K$.
  } 
  \label{fig: K selection}
\end{figure*}

We prepare the PTB \citep{marcusetal1993building} for English and the SPMRL dataset
\citep{seddahetal2013overview} for eight other languages.
We adopt the standard split of each dataset to divide it
into development and test sets.
For preprocessing, we follow the setting in \citet{kimetal2019compound, kimetal2019unsupervised}. 

We run our ranking algorithm on the development set to select the syntax-related heads and the ensemble parsing algorithm on the test set.
We only use the raw sentences in the development set, without any syntactic annotations.
We average all configurations both for ranking ($f$, $s_{comp}$ and $s_{cross}$) and parsing ($f$ and $s_{comp}$); hence we do not tune any hyperparameters for our algorithm.
For $K$ selection, we experiment with fixed top-$K$ (i.e., top-30) and dynamically searching the best $K$ described in Section~\ref{subsec: k selection}, dubbed dynamic~$K$.
We report the unlabeled sentence-level $F_1$ score to evaluate the extent to which the induced trees resemble the corresponding gold standard trees.

For neural PCFG training, we modify some details but keep most of the model configurations of \citet{kimetal2019compound}; we refer readers to the original paper for more information. 
We train the models on longer sentences for more epochs.
Specifically, we train on sentences of length up to 30 in the first epoch, and increase this length limit by five until the length reaches~80.
We train for 30 epochs and use a learning rate scheduler.


\begingroup
\setlength{\tabcolsep}{4pt}
\begin{table}[tb]\centering
  \resizebox{1.02\linewidth}{!}{
  \begin{tabular}{l | c | c c | c | c c c}
    \toprule
    Model					& Top-down     &  \multicolumn{3}{c|}{Chart-based}  & \multicolumn{3}{c }{Our ranking-based} \\
    \midrule
    \multirow{3}{*}{Configuration} & Single & Single   & Top     & Top    & Top & Dynamic & Full \\
                           & /Layer$^\dagger$ & /Layer$^\dagger$ & -$K$   &-$K$$^\ddagger$    & -$K$ & $K$    & heads \\
                            \cmidrule{2-8}
                           & \multicolumn{3}{c |}{w/ dev trees} & \multicolumn{4}{ c}{w/o dev trees} \\
    \midrule
    BERT-base-cased & 32.6 & 37.5 & \textbf{42.7} & 29.3 & 34.8 & \textbf{37.1} & 35.8 \\ 
    BERT-large-cased & 36.7 & 41.5 & \textbf{44.6} & 21.5 & 36.1 & \textbf{38.7} & 33.2 \\
    XLNet-base-cased & 39.0 & 40.5 & \textbf{46.4} & 38.4 & 41.2 & \textbf{42.7} & 42.4 \\
    XLNet-large-cased & 37.3 & 39.7 & \textbf{46.4} & 34.1 & 40.6 & 41.1 & \textbf{41.2} \\
    RoBERTa-base & 38.0 & 41.0 & \textbf{45.0} & 35.9 & 41.7 & \textbf{42.1} & 39.6 \\
    RoBERTa-large & 33.8 & 38.6 & \textbf{42.8} & 30.2 & 33.1 & \textbf{37.5} & 35.7 \\
    GPT2 & 35.4 & 34.5 & \textbf{38.5} & 21.9 & 26.1 & \textbf{27.2} & 26.1 \\
    GPT2-medium & 37.8 & 38.5 & \textbf{39.8} & 19.4 & \textbf{29.1} & \textbf{29.1} & 27.2 \\
    \midrule
    AVG & 36.3 & 39.0 & \textbf{43.3} & 28.8 & 35.3 & \textbf{36.9} & 35.1 \\
    AVG w/o GPT2~* & 36.2 & 39.8 & \textbf{44.7} & 31.6 & 37.9 & \textbf{39.8} & 38.0 \\
    \bottomrule
  \end{tabular}}
  \caption{
    Unlabeled sentence-level parsing $F_1$ scores on the English PTB test set. 
    $\dagger$: the best results of the top single head and layer-wise ensemble. 
    $\ddagger$: directly applying the chart-based parser for ranking (no development set trees) and ensembling the top-$K$ heads for parsing.
    *: average $F_1$ scores without GPT2 and GPT2-medium.
    Bold figures highlight the best scores for the two different groups: with and without development trees.
    }
  \label{tab: PTB f1}
  \end{table}
  \endgroup

\begingroup
\setlength{\tabcolsep}{4pt}
\begin{table}[tb]\centering
  \resizebox{\linewidth}{!}{
  \begin{tabular}{l | c | cccccc}
  \toprule
  Model &$F_1$ &SBAR &NP &VP &PP &ADJP &ADVP \\
  \midrule
  Balanced  &18.5 &7 &27 &8 &18 &27 &25 \\
  Left branching  &8.7 &5 &11 &0 &5 &2 &8 \\
  Right branching  &39.4 &\textbf{68} &24 &\textbf{71} &42 &27 &38 \\
  \midrule
  BERT-base-cased &37.1 &36 &49 &30 &42 &40 &69 \\
  BERT-large-cased &38.7 &38 &50 &30 &46 &42 &\textbf{72} \\
  XLNet-base-cased &\textbf{42.7} &45 &\textbf{58} &31 &46 &46 &\textbf{72} \\
  XLNet-large-cased &41.1 &44 &54 &30 &42 &\textbf{48} &64 \\
  RoBERTa-base &42.1 &38 &\textbf{58} &31 &\textbf{47} &42 &71 \\
  RoBERTa-large &37.5 &35 &53 &29 &33 &36 &54 \\
  \bottomrule
  \end{tabular}}
  \caption{
    Unlabeled parsing scores and recall scores on six constituency tags of trivial baseline parse trees as well as ones achieved by our parser using dynamic $K$ on different PLMs. }
  \label{tab: PTB tags}
  \end{table}
  \endgroup

\subsection{Results on the English PTB}

We first evaluate our ranking-based zero-shot parser on the English PTB dataset.
We apply our methods to four different PLMs for English: BERT \citep{devlinetal2019bert}, XLNet \citep{yang2019xlnet}, RoBERTa \citep{liu2019roberta}, and GPT2 \citep{radford2019language}.\footnote{
We follow previous work \citep{Kim2020Are, kim2020multilingual} in using two variants for each PLM, where the X-base variants consist of 12 layers, 12 attention heads, and 768 hidden dimensions, while the X-large ones have 24 layers, 16 heads, and 1024 dimensions. 
With regard to GPT2, the GPT2 model corresponds to X-base while GPT2-medium to X-large.
}

Table~\ref{tab: PTB f1} shows the unlabeled $F_1$ scores for our ranking-based zero-shot parser
as well as for previous zero-shot parsers in two settings, with and without an annotated development set. 
We employ the chart-based parser in a setting without development trees, where Eqs.~(\ref{eq: recursive span}) and (\ref{eq: recursive split}) are used for ranking and ensembling the top-$K$ (i.e., top-30) heads.
Compared to our method under the same configuration, its poor performance confirms the effectiveness of our ranking algorithm.

With respect to the $K$ selection, our dynamic $K$ method beats both fixed top-30 and full heads.
Surprisingly, using all attention heads for ensemble parsing yields nearly the same performance as using top-30 heads. 
This suggests that although our ranking algorithm filters out some noisy heads, it is still not perfect.
On the other hand, the ensemble parsing method is robust to noisy heads when full attention heads are used. 
Figure~\ref{fig: K selection} shows how the ensemble parsing performance changes given different $K$ selection.
We can identify a roughly concave shape of the parsing performance curve, which indicates why our ranking algorithm works.
Interestingly, the parsing performance does not drop too much when $K$ reaches the maximum for XLNet.
We conjecture that syntactic knowledge is more broadly distributed across heads in XLNet.

Our ranking-based parser performs badly on GPT2 and GPT2-medium, which is not unexpected. 
Unlike other PLMs, models in the GPT2 category are auto-regressive language models, whose attention matrix is strictly lower triangular.
It makes it hard for our ranking algorithm to work properly.
But for top-down and chart-based zero-shot parsers, tuning against an annotated development set can alleviate this problem.
We focus on BERT, XLNet and RoBERTa and only evaluate these three models in the rest of our experiments.
Except for GPT2 variants, our parser with dynamic $K$ outperforms the top-down parser in all cases. 
On average (without GPT2 variants), even though our parser only requires raw sentence data, it still matches the chart-based parser with the top single head or layer-wise ensemble.
To explore the limit of the chart-based parser, we also present the results by selecting the top-$K$ (i.e., top-20) heads using the annotated development set \citep{kim2020multilingual}.
\footnote{Selecting heads against a development set ensures the quality of high ranking heads; top-20 heads are optimal in this setting \citep{kim2020multilingual}, unlike top-30 in our setting.}
Note that in this setting, the best configuration, i.e., the combination of $g$, $f$ and $s_{comp}$ as well as $K$ are selected against the development set.
This setting serves as an upper bound of the chart-based zero-shot parsing and largely outperforms our ranking-based method.

\begingroup
\setlength{\tabcolsep}{8pt}
\begin{table*}[!htp]\centering
  \resizebox{.95\linewidth}{!}{
  \begin{tabular}{c | l | ccccccccc | c}
    \toprule
    \multicolumn{2}{c |}{Model} & English & Basque & French & German & Hebrew & Hungarian & Korean & Polish & Swedish & AVG \\
    \midrule
    \multicolumn{12}{c}{Trivial baselines } \\
    \midrule
    \multicolumn{2}{l |}{Balanced} & 18.5 & 24.4 & 12.9 & 15.2 & 18.1 & 14.0 & 20.4 & 26.1 & 13.3 & 18.1 \\
    \multicolumn{2}{l |}{Left branching} & 8.7 & 14.8 & 5.4 & 14.1 & 7.7 & 10.6 & 16.5 & 28.7 & 7.6 & 12.7 \\
    \multicolumn{2}{l |}{Right branching} & 39.4 & 22.4 & 1.3 & 3.0 & 0.0 & 0.0 & 21.1 & 0.7 & 1.7 & 10.0 \\
    \midrule
    
    \multirow{14}{*}{\rotatebox[origin=c]{90}{w/ dev trees}} & \multicolumn{11}{c}{ Chart-based (Single/Layer) $^{\dagger}$ } \\
    \cmidrule[0.5pt]{2-12}
    & M-BERT &41.2 &38.1 &30.6 &32.1 &31.9 &30.4 &46.4 &43.5 &27.5 &35.7 \\
    & XLM &43.0 &35.3 &35.6 &41.6 &39.9 &34.5 &35.7 &51.7 &33.7 &39.0 \\
    & XLM-R &44.4 &40.4 &31.0 &32.8 &34.1 &32.4 &47.5 &44.7 &29.2 &37.4 \\
    & XLM-R-large &40.8 &36.5 &26.4 &30.2 &32.1 &26.8 &45.6 &47.9 &25.8 &34.7 \\
    \cmidrule[0.3pt]{2-12}
    & AVG & 42.4 &37.6 &30.9 &34.2 &34.5 &31.0 &43.8 &46.9 &29.1 &
    36.7 \\
    \cmidrule[0.5pt]{2-12}

    & \multicolumn{11}{c}{Chart-based (top-$K$) $^{\dagger}$ } \\
    \cmidrule[0.5pt]{2-12}
    & M-BERT &45.0 &41.2 &35.9 &35.9 &37.8 &33.2 &47.6 &51.1 &32.6 &40.0 \\
    & XLM &\textbf{47.7} &41.3 &\textbf{36.7} &\textbf{43.8} &\textbf{41.0} &36.3 &35.7 &\textbf{58.5} &\textbf{36.5} &\textbf{41.9} \\
    & XLM-R &47.0 &\textbf{42.2} &35.8 &37.7 &40.1 &\textbf{36.6} &\textbf{51.0} &52.7 &32.9 &41.8 \\
    & XLM-R-large &45.1 &40.2 &29.7 &37.1 &36.2 &31.0 &46.9 &47.9 &27.8 &38.0 \\
    \cmidrule[0.3pt]{2-12}
    & AVG &46.2 &41.2 &34.5 &38.6 &38.8 &34.3 &45.3 &52.6 &32.5 &40.4 \\
    \midrule
    
    \multirow{6}{*}{\rotatebox[origin=c]{90}{w/o dev trees}}   & \multicolumn{11}{c}{ Crosslingual ranking-based (Dynamic $K$) $^{\ddagger}$ } \\
    \cmidrule[0.5pt]{2-12}
    & M-BERT & 40.7 &\textbf{38.2} &31.0 &31.0 &29.0 &27.1 &43.3 &30.7 &25.8 &33.0 \\
    & XLM & 44.9 &26.6 &\textbf{35.8} &\textbf{39.7} &\textbf{39.6} &32.9 &28.0 &\textbf{50.1} &\textbf{34.1} &36.9 \\
    & XLM-R & \textbf{45.5} &\textbf{38.2} &34.0 &35.5 &36.7 &\textbf{33.5} &\textbf{45.2} &39.4 &29.9 &\textbf{37.6} \\
    & XLM-R-large & 41.0 &37.9 &28.0 &28.0 &31.3 &24.6 &44.4 &32.2 &24.9 &32.5 \\
    \cmidrule[0.3pt]{2-12}
    & AVG &43.0 &34.7 &32.4 &33.5 &35.0 &29.8 &40.4 &39.2 &29.2 &35.3 \\
    
    \bottomrule
    \end{tabular}}
    \caption{
      Parsing results on nine languages with multilingual PLMs. 
      $^{\dagger}$: attention heads are selected on the development trees in the target language. 
      $^{\ddagger}$: attention heads are selected on raw sentences in English.
      Bold figures highlight the best scores for the two different groups: with and without development trees.
      }
    \label{tab: multi}
\end{table*}
\endgroup

Table~\ref{tab: PTB tags} presents the parsing scores as well as recall scores on different constituents of trivial baselines and our parser.
It indicates that trees induced from XLNet-base-cased, XLNet-large-cased and RoBERTa-base can outperform the right-branching baseline without resembling it.
This confirms that PLMs can produce non-trivial parse trees.
Large gains on NP, ADJP and ADVP compared to the right branching baseline show that PLMs can better identify such constituents.

\subsection{Results for Languages other than English}

Low-resource language parsing is one of the main motivations for the development of unsupervised parsing algorithms, which makes a multilingual setting ideal for evaluation.
Multilingual PLMs are attractive in this setting because they are trained to process over one hundred languages in a language-agnostic manner. 
\citet{kim2020multilingual} has investigated the zero-shot parsing capability of multilingual PLMs assuming that a small annotated development set is available.
Here, by taking advantage of our ranking-based parsing algorithm, we use a more radical crosslingual setting.
We rank attention heads only on sentences in English and directly apply the parser to eight other languages.
We follow \citet{kim2020multilingual} and use four multilingual PLMs: 
a multilingual version of the BERT-base model (M-BERT, \citealt{devlinetal2019bert}), the XLM model \citep{Conneau2019xlm}, the XLM-R and XLM-R-large models \citep{conneau2019unsupervised}.
Each multilingual PLM differs in architecture and pre-training data, and we refer readers to the original papers for more details.

In Table~\ref{tab: multi}, our crosslingual parser outperforms the trivial baselines in all cases by a large margin. 
Compared with the chart-based parser with the top head or layer-wise ensemble, our crosslingual parser can match the performance on five out of nine languages.
Among four model variants, XLM-R and XLM-R-large have identical training settings and pre-training data, and so form a controlled experiment.
By directly comparing XLM-R and XLM-R-large, we conjecture that, as the capacity of the PLM scales, the model has more of a chance to learn separate hidden spaces for different languages. 
This is consistent with a recent study on multilingual BERT \citep{dufter2020identifying} showing that underparameterization is one of the main factors that contribute to multilinguality.
Again, our method lags behind the chart-base zero-shot parser with a top-$K$ ensemble.
More experimental results including using target language for head selection in our method can be found in Appendix~\ref{appendix: full multi}.

\subsection{Grammar Analysis}

By not relying on an annotated development set, we have an unbiased way of investigating the tree structures as well as the grammars that are inherent in PLMs.
Specifically, we first parse the raw sentences using our ranking-based parser described in Section~\ref{subsec: ranking-parser} and then train a neural PCFG given the induced trees using the method in Section~\ref{subsec: learn grammar}.
We conduct our experiments on the English PTB and evaluate how the learned grammar resembles PTB syntax in a quantitative way on preterminals (PoS tags) and production rules.
We visualize the alignment of preterminals and nonterminals of the learned grammar and the gold labels in Appendix~\ref{appendix: T NT alignment} as a qualitative study.
We also showcase parse trees of the learned grammar to get a glimpse of some distinctive characteristics of the learned grammar in Appendix~\ref{appendix: tree samples}.
For brevity, we refer to a neural PCFG learned from trees induced of a PLM as PCFG\subsetting{PLM} and to a neural PCFG learned from the gold parse trees as PCFG\subsetting{Gold}.

%

\begingroup
\setlength{\tabcolsep}{10pt}
\begin{table}[tb]\centering
  \resizebox{0.95\linewidth}{!}{
    \begin{tabular}{l | c | c | c}
      \toprule
      \multirow{2}{*}{Trees} & Preterminal & Rule & Parsing  \\
       & Acc$^\dagger$ & Acc$^\ddagger$ & $F_1$ \\
      \midrule
  Gold* &66.1 &\textbf{46.2} & - \\
  BERT-base-cased &64.4 &24.8 &37.1 \\
  BERT-large-cased &64.0 &22.3 &38.7 \\
  XLNet-base-cased &\textbf{67.7} &26.1 &42.7 \\
  XLNet-large-cased &65.8 &27.3 &41.1 \\
  RoBERTa-base &65.7 &27.2 &42.1 \\
  RoBERTa-large &62.4 &25.1 &37.5 \\
      \bottomrule
    \end{tabular}}
    \caption{Preterminal (PoS tag) and production rule accuracies of PCFG\subsetting{PLM} and PCFG\subsetting{Gold} on the entire PTB. 
    $\dagger$: PoS tagging accuracy using the many-to-one mapping \citep{johnson2007doesn}. 
    $\ddagger$: production rule accuracy where anonymized nonterminals and preterminals are mapped to the gold tags using the many-to-one mapping.
    *: PCFG\subsetting{Gold}.
    }
    \label{tab: POS and rule acc}
  \end{table}
\endgroup

In Table~\ref{tab: POS and rule acc}, we report preterminal (unsupervised PoS tagging) accuracies and production rule accuracies of PCFG\subsetting{PLM} and PCFG\subsetting{Gold} on the corpus level.
For preterminal evaluation, we map the anonymized preterminals to gold PoS tags using many-to-one (M-1) mapping \citep{johnson2007doesn}, where each anonymized preterminal is
matched onto the gold PoS tag with which it shares the most tokens. 
For production rule evaluation, we map both nonterminals and preterminals to gold tags using M-1 mapping to get the binary production rules.\footnote{
  For the gold annotations, we drop all unary rules.
  For $n$-ary rules ($n > 2$), we convert them to binary rules by right branching and propagating the parent tag. For example, a $n$-ary rule $A \to B\ C\ D$ yields $A \to B\ A$ and $A \to C\ D$.
}
We find that all PCFG\subsetting{PLM} grammars except for PCFG\subsetting{RoBERTa-large} outperform a discrete HMM baseline (62.7, \citealt{he2018unsupervised}) but are far from the state of the art for neural grammar induction (80.8, \citealt{he2018unsupervised}). 
All PCFG\subsetting{PLM} produce similar accuracies on preterminals as PCFG\subsetting{Gold}.
However, for the production rules, PCFG\subsetting{PLM} lags behind PCFG\subsetting{Gold} by a large margin.
This makes sense as presumably the tree structures heavily affect nonterminal learning.
We also present the parsing $F_1$ scores of corresponding trees against the gold trees in Table~\ref{tab: POS and rule acc} for comparison.
We observe that for all PCFG\subsetting{PLM}, both preterminal accuracies and production rule accuracies correlate well with the parsing $F_1$ scores of the corresponding trees.

\section{Conclusion}

In this paper, we set out to analyze the syntactic knowledge learned
by transformer-based pre-trained language models. In contrast to
previous work relying on test suites and probes, we proposed to use a
zero-shot unsupervised parsing approach. This approach is able to
parse sentences by ranking the attention heads of the PLM and
ensembling them. Our approach is able to completely do away with a
development set annotated with syntactic structures, which makes it
ideal in a strictly unsupervised setting, e.g., for low resource
languages.  We evaluated our method against previous methods on nine
languages.  When development sets are available for previous methods,
our method can match them or produce competitive results if they use
the top single head or layer-wise ensembling of attention heads, but
lags behind them if they ensemble the top-$K$ heads.  Furthermore, we
present an analysis of the grammars learned by our approach: we use
the induced trees to train a neural PCFG and evaluate the pre-terminal
and non-terminal symbols of that grammar. In future work, we will
develop further methods for analyzing the resulting grammar rules.
Another avenue for follow-up research is to use our method to
determine how the syntactic structures inherent in PLMs change when
these models are fine-tuned on a specific task.

\section*{Acknowledgments}
We thank the reviewers for their valuable suggestions regarding this work.


\bibliography{anthology,aacl-ijcnlp2020}
\bibliographystyle{acl_natbib}

\clearpage
\newpage

\appendix
\section{Appendix}

\subsection{More Results on Languages other than English}
\label{appendix: full multi}

We present a comprehensive analysis of the chart-based parser and our ranking-based parser on the multilingual setting.
In addition to Table~\ref{tab: multi}, for our method, we conduct experiments using target language for head selection with both Top-$K$ (i.e., top-30) ensemble and dynamic $K$ ensemble.

In Table~\ref{tab: full multi}, we find that our ranking-based parser with Top-$K$ ensemble performs slightly better than that using dynamic $K$.
In contrast to the superiority of dynamic $K$ on English PLMs in Table~\ref{tab: PTB f1}, multilingual PLMs produce similar parsing performance with a \textit{lazy} top-30 ensemble.
We conjecture that there could be no clear concave pattern (like Figure~\ref{fig: K selection}) in the relation of $K$ and parsing performance in this crosslingual setting.

We also experimented with another setting for our ranking-based parser: selecting attention heads based on the sentences in the target language.
Interestingly, we observe a considerable parsing performance drop on both top-$K$ and dynamic $K$ ensemble.
We suspect that our chart-based ranking algorithm (e.g.,~the inherent context free grammar assumption) does not work equally well in all languages, at least for the annotation scheme provided by the SPMRL dataset.
In this scenario, using English for head selection has a better chance to capture syntax-related attention heads.
Again, as we discussed before, using annotated trees in the target language can always ensure the quality of selected top-$K$ heads.

\begingroup
\begin{table*}[!htp]\centering
  \resizebox{1.0\linewidth}{!}{
  \begin{tabular}{c | l | ccccccccc | c}
    \toprule
    \multicolumn{2}{c |}{Language} & English & Basque & French & German & Hebrew & Hungarian & Korean & Polish & Swedish & AVG \\
    \midrule
    \multicolumn{12}{c}{ Trivial baselines } \\
    \midrule
    \multicolumn{2}{l |}{Balanced} & 18.5 & 24.4 & 12.9 & 15.2 & 18.1 & 14.0 & 20.4 & 26.1 & 13.3 & 18.1 \\
    \multicolumn{2}{l |}{Left branching} & 8.7 & 14.8 & 5.4 & 14.1 & 7.7 & 10.6 & 16.5 & 28.7 & 7.6 & 12.7 \\
    \multicolumn{2}{l |}{Right branching} & 39.4 & 22.4 & 1.3 & 3.0 & 0.0 & 0.0 & 21.1 & 0.7 & 1.7 & 10.0 \\
    \midrule

    \multirow{24}{*}{\rotatebox[origin=c]{90}{Target language for head selection}} & \multicolumn{11}{c}{ Chart-based (Single/Layer) $^{\dagger}$ } \\
    \cmidrule[0.5pt]{2-12}
    & M-BERT &41.2 &38.1 &30.6 &32.1 &31.9 &30.4 &46.4 &43.5 &27.5 &35.7 \\
    & XLM &43.0 &35.3 &35.6 &41.6 &39.9 &34.5 &35.7 &51.7 &33.7 &39.0 \\
    & XLM-R &44.4 &40.4 &31.0 &32.8 &34.1 &32.4 &47.5 &44.7 &29.2 &37.4 \\
    & XLM-R-large &40.8 &36.5 &26.4 &30.2 &32.1 &26.8 &45.6 &47.9 &25.8 &34.7 \\
    \cmidrule[0.3pt]{2-12}
    & AVG & 42.4 &37.6 &30.9 &34.2 &34.5 &31.0 &43.8 &46.9 &29.1 &
    36.7 \\
    \cmidrule[0.5pt]{2-12}

    & \multicolumn{11}{c}{ Chart-based (Top-$K$) $^{\dagger}$ } \\
    \cmidrule[0.5pt]{2-12}
    & M-BERT &45.0 &41.2 &35.9 &35.9 &37.8 &33.2 &47.6 &51.1 &32.6 &40.0 \\
    & XLM &\textbf{47.7} &41.3 &\textbf{36.7} &\textbf{43.8} &\textbf{41.0} &36.3 &35.7 &\textbf{58.5} &\textbf{36.5} &\textbf{41.9} \\
    & XLM-R &47.0 &\textbf{42.2} &35.8 &37.7 &40.1 &\textbf{36.6} &\textbf{51.0} &52.7 &32.9 &41.8 \\
    & XLM-R-large &45.1 &40.2 &29.7 &37.1 &36.2 &31.0 &46.9 &47.9 &27.8 &38.0 \\
    \cmidrule[0.3pt]{2-12}
    & AVG &46.2 &41.2 &34.5 &38.6 &38.8 &34.3 &45.3 &52.6 &32.5 &40.4 \\
    \cmidrule[0.5pt]{2-12}

    & \multicolumn{11}{c}{ Ranking-based (Top-$K$) $^{\ddagger}$ } \\
    \cmidrule[0.5pt]{2-12}
    & M-BERT &41.5 &38.9 &33.9 &30.2 &36.3 &30.9 &39.0 &18.4 &26.3 &31.7 \\
    & XLM &44.6 &21.0 &29.8 &39.2 &30.5 &25.2 &23.8 &55.2 &30.3 &31.9 \\
    & XLM-R &44.8 &36.0 &34.1 &31.8 &36.4 &32.5 &40.3 &29.6 &26.7 &33.4 \\
    & XLM-R-large &41.1 &36.8 &30.3 &26.8 &33.4 &24.9 &37.4 &17.5 &26.3 &29.2 \\
    \cmidrule[0.3pt]{2-12}
    & AVG & 43.0 &33.2 &32.0 &32.0 &34.2 &28.4 &35.1 &30.2 &27.4 &31.6 \\
    \cmidrule[0.5pt]{2-12}

    & \multicolumn{11}{c}{ Ranking-based (Dynamic $K$) $^{\ddagger}$ } \\
    \cmidrule[0.5pt]{2-12}
    & M-BERT &40.7 &39.1 &28.4 &25.5 &26.9 &31.2 &41.3 &22.2 &21.3 &29.5 \\
    & XLM &44.9 &20.8 &29.9 &40.3 &34.4 &27.7 &23.6 &55.1 &31.2 &32.9 \\
    & XLM-R &45.5 &37.3 &30.7 &31.5 &31.8 &34.1 &40.8 &36.0 &27.4 &33.7 \\
    & XLM-R-large &41.0 &36.5 &29.0 &30.1 &32.6 &25.3 &43.9 &30.0 &25.5 &31.6 \\
    \cmidrule[0.3pt]{2-12}
    & AVG &43.0 &33.4 &29.5 &31.9 &31.4 &29.6 &37.4 &35.8 &26.4 &31.9 \\
    \midrule

    \multirow{12}{*}{\rotatebox[origin=c]{90}{English for head selection}} & \multicolumn{11}{c}{ Crosslingual ranking-based (Top-$K$) $^{\ddagger}$ } \\
    \cmidrule[0.5pt]{2-12}
    & M-BERT &- &37.9 &33.4 &31.2 &31.5 &29.4 &45.3 &33.4 &27.2 &34.5 \\
    & XLM &- &25.9 &34.4 &39.2 &39.5 &31.9 &27.5 &\textbf{50.4} &\textbf{34.2} &36.4 \\
    & XLM-R &- &37.9 &33.9 &35.1 &36.8 &33.3 &44.7 &39.7 &30.3 &37.4 \\
    & XLM-R-large &- &35.7 &28.5 &28.5 &34.7 &25.5 &44.5 &36.9 &27.1 &33.6 \\
    \cmidrule[0.3pt]{2-12}
    & AVG &- &34.3 &32.6 &33.5 &35.6 &30.0 &40.5 &40.1 &29.7 &35.5 \\
    \cmidrule[0.5pt]{2-12}

    & \multicolumn{11}{c}{ Crosslingual ranking-based (Dynamic $K$) $^{\ddagger}$ } \\
    \cmidrule[0.5pt]{2-12}
    & M-BERT & - &\textbf{38.2} &31.0 &31.0 &29.0 &27.1 &43.3 &30.7 &25.8 &33.0 \\
    & XLM & - &26.6 &\textbf{35.8} &\textbf{39.7} &\textbf{39.6} &32.9 &28.0 &50.1 &34.1 &36.9 \\
    & XLM-R & - &\textbf{38.2} &34.0 &35.5 &36.7 &\textbf{33.5} &\textbf{45.2} &39.4 &29.9 &\textbf{37.6} \\
    & XLM-R-large & - &37.9 &28.0 &28.0 &31.3 &24.6 &44.4 &32.2 &24.9 &32.5 \\
    \cmidrule[0.3pt]{2-12}
    & AVG &- &34.7 &32.4 &33.5 &35.0 &29.8 &40.4 &39.2 &29.2 &35.3 \\

    \bottomrule
    \end{tabular}}
    \caption{
      Parsing results on nine languages with multilingual PLMs. 
      Except for the trivial baselines, all experimental results are divided into two groups: using target language for head selection and using English for head selection (crosslingual).
      $^{\dagger}$: results of the best configurations of $f$, $g$, $s_{comp}$ and $K$ are decided on an annotated development set. 
      $^{\ddagger}$: results where only raw sentences are required.
      For top-$K$, 20 is used for chart-based and 30 is used for our ranking-based.
      Bold figures highlight the best scores for the two different groups: using target language and English for head selection.
      }
    \label{tab: full multi}
\end{table*}
\endgroup

\subsection{Visualization of the Alignment for Internal Tags}
\label{appendix: T NT alignment}

\begin{figure*}[!htp]
  \centering
  \begin{subfigure}[b]{0.24\textwidth}
      \centering
      \includegraphics[width=\textwidth]{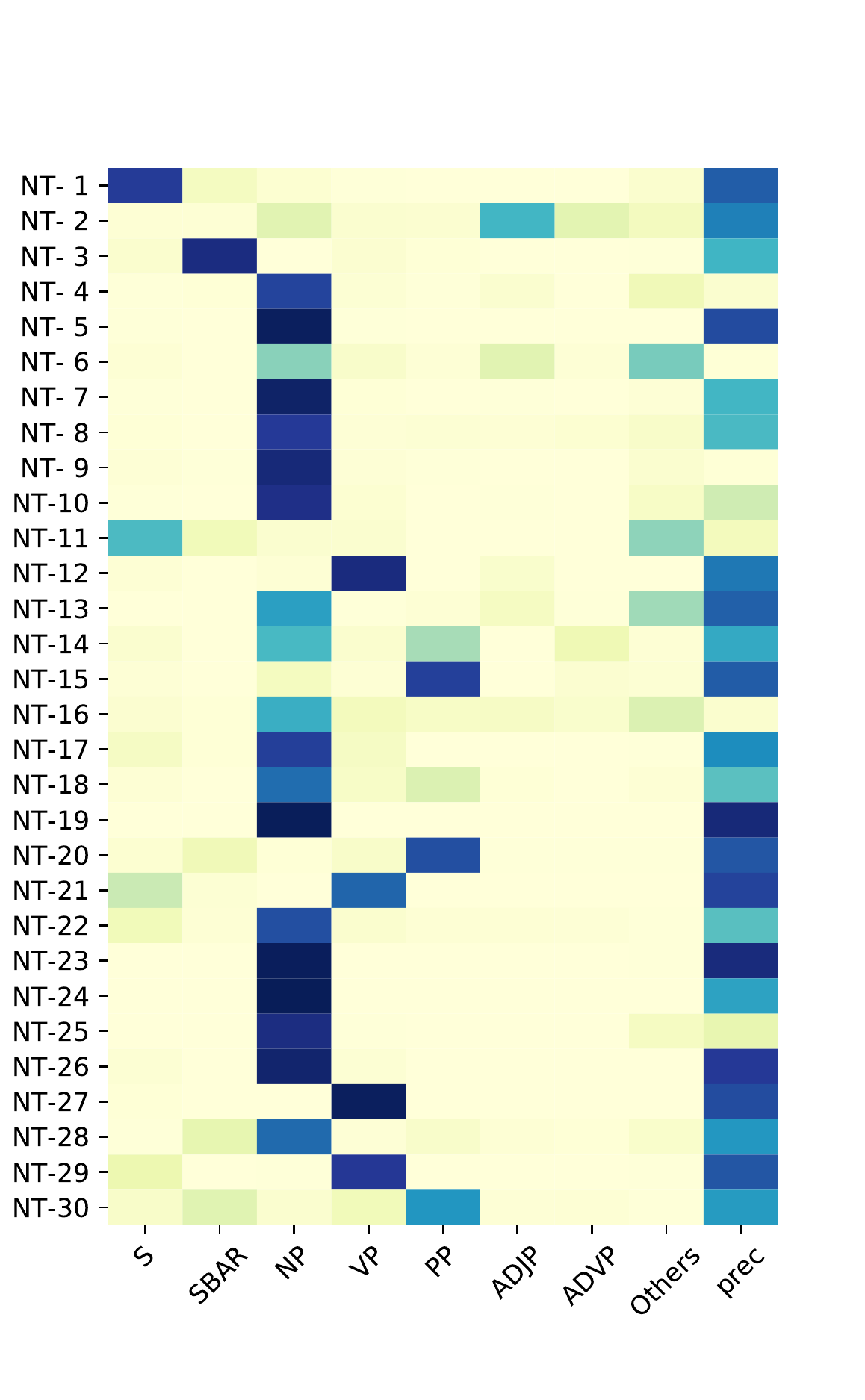}
      \caption{PCFG$_{\text{Gold}}$}    
  \end{subfigure} \hfil
  \begin{subfigure}[b]{0.24\textwidth}
    \centering
    \includegraphics[width=\textwidth]{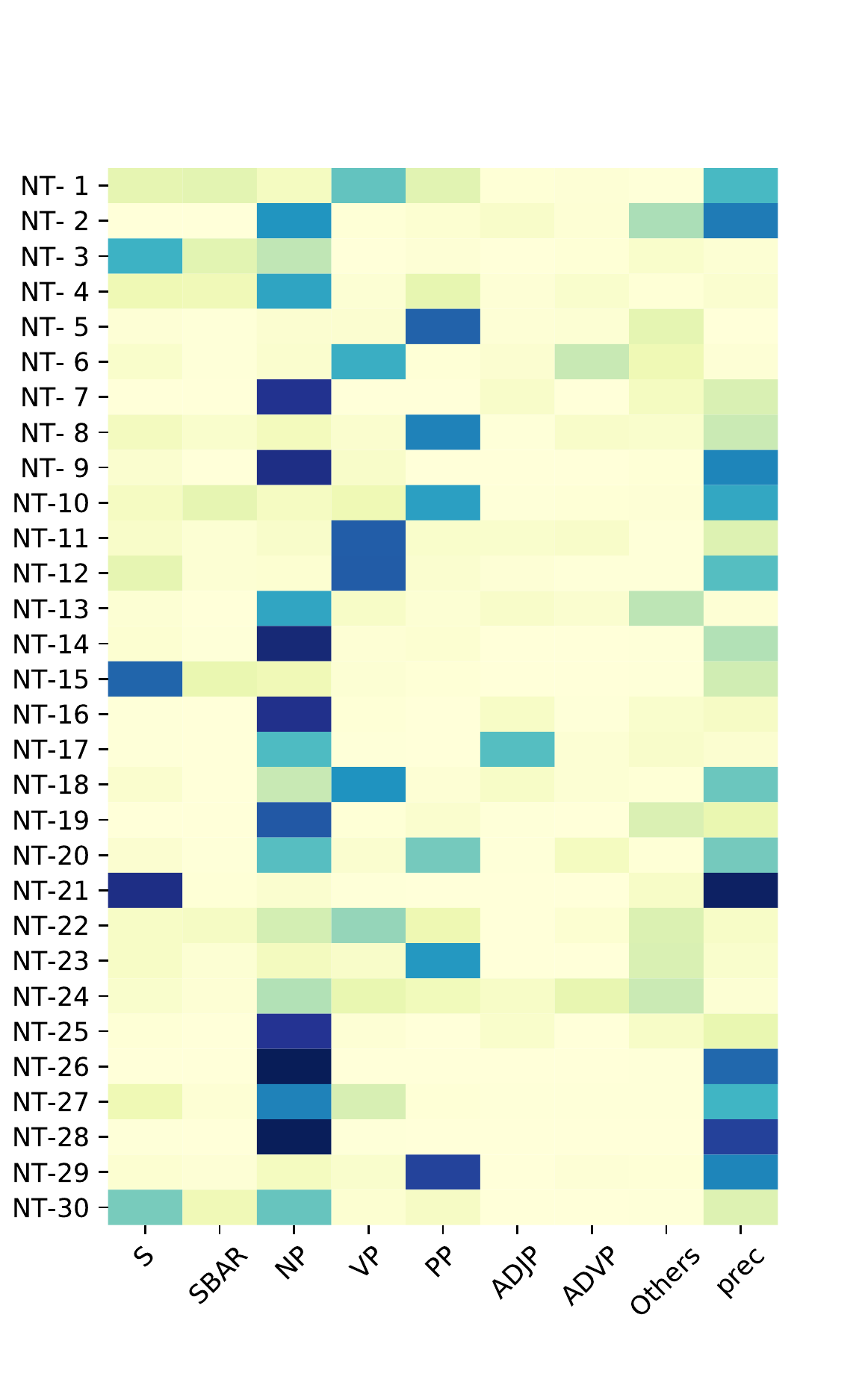}
    \caption{PCFG$_{\text{BERT-base-cased}}$}    
  \end{subfigure} \hfil
  \begin{subfigure}[b]{0.24\textwidth}   
      \centering 
      \includegraphics[width=\textwidth]{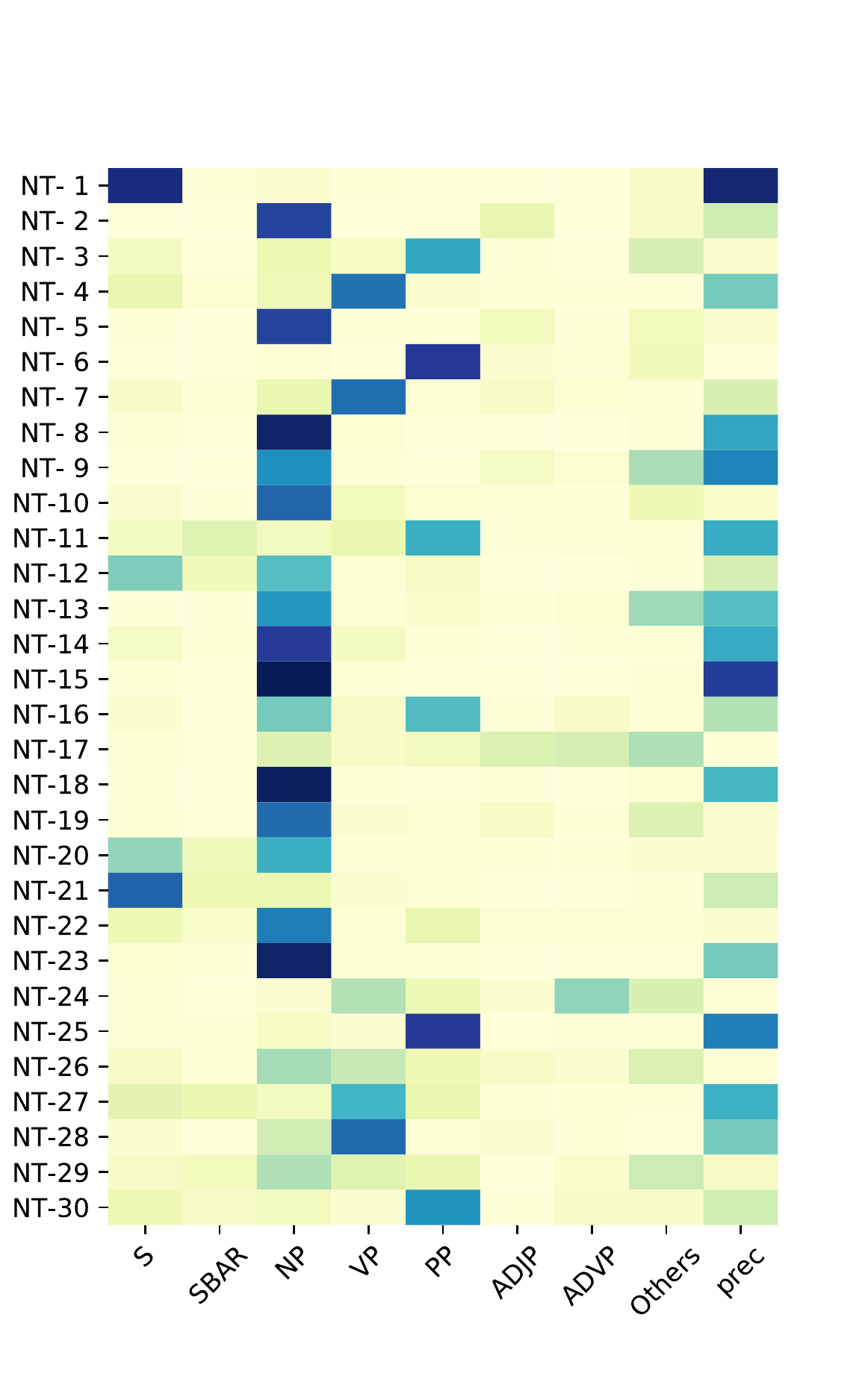}
      \caption{PCFG$_{\text{BERT-large-cased}}$}    
  \end{subfigure} \hfil
  \begin{subfigure}[b]{0.24\textwidth}   
    \centering 
    \includegraphics[width=\textwidth]{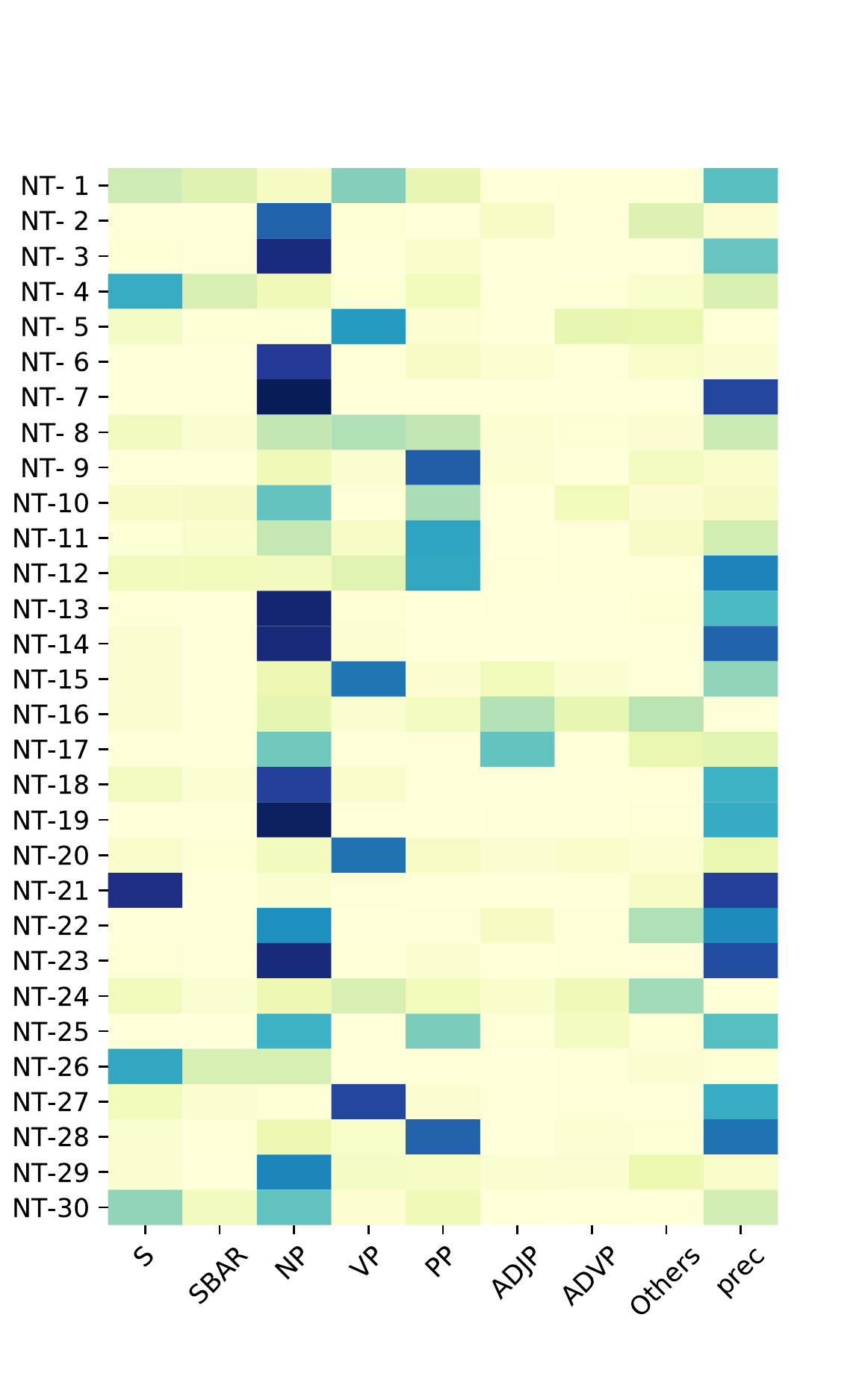}
    \caption{PCFG$_{\text{XLNet-base-cased}}$}    
  \end{subfigure}

  \begin{subfigure}[b]{0.24\textwidth}   
    \centering 
    \includegraphics[width=\textwidth]{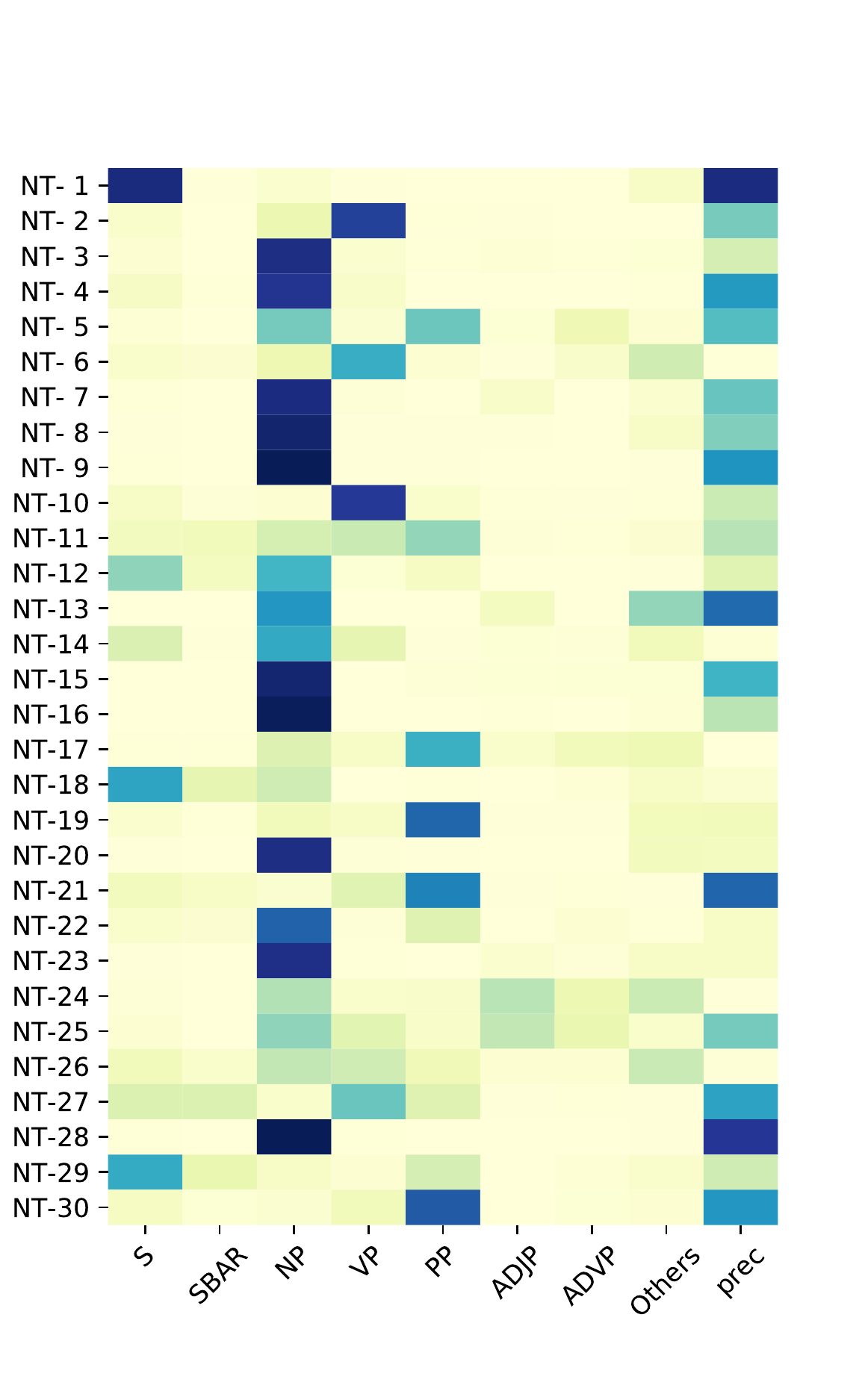}
    \caption{PCFG$_{\text{XLNet-large-cased}}$}    
  \end{subfigure} \hfil
  \begin{subfigure}[b]{0.24\textwidth}   
    \centering 
    \includegraphics[width=\textwidth]{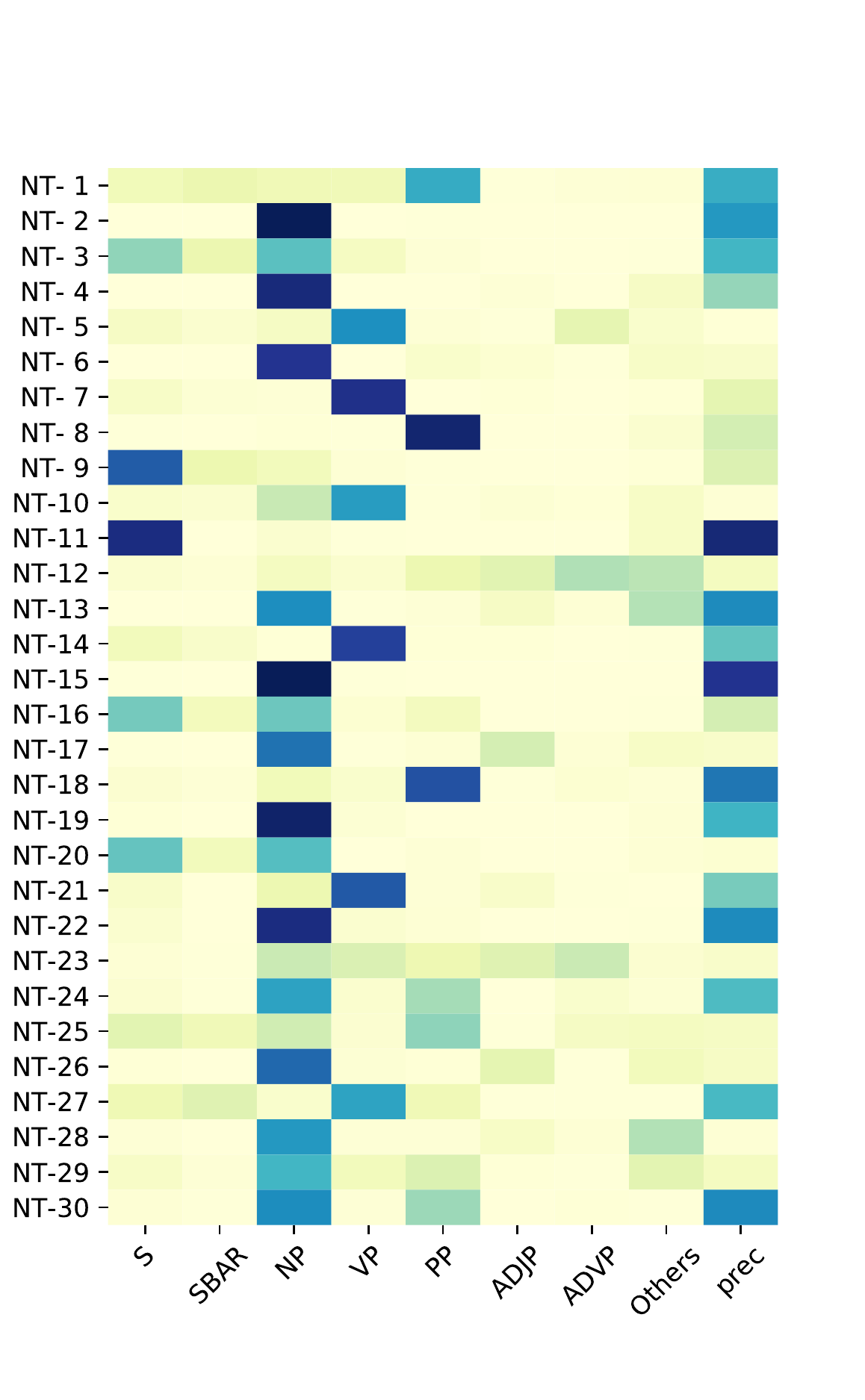}
    \caption{PCFG$_{\text{RoBERTa-base}}$}    
  \end{subfigure} \hfil
  \begin{subfigure}[b]{0.24\textwidth}   
    \centering 
    \includegraphics[width=\textwidth]{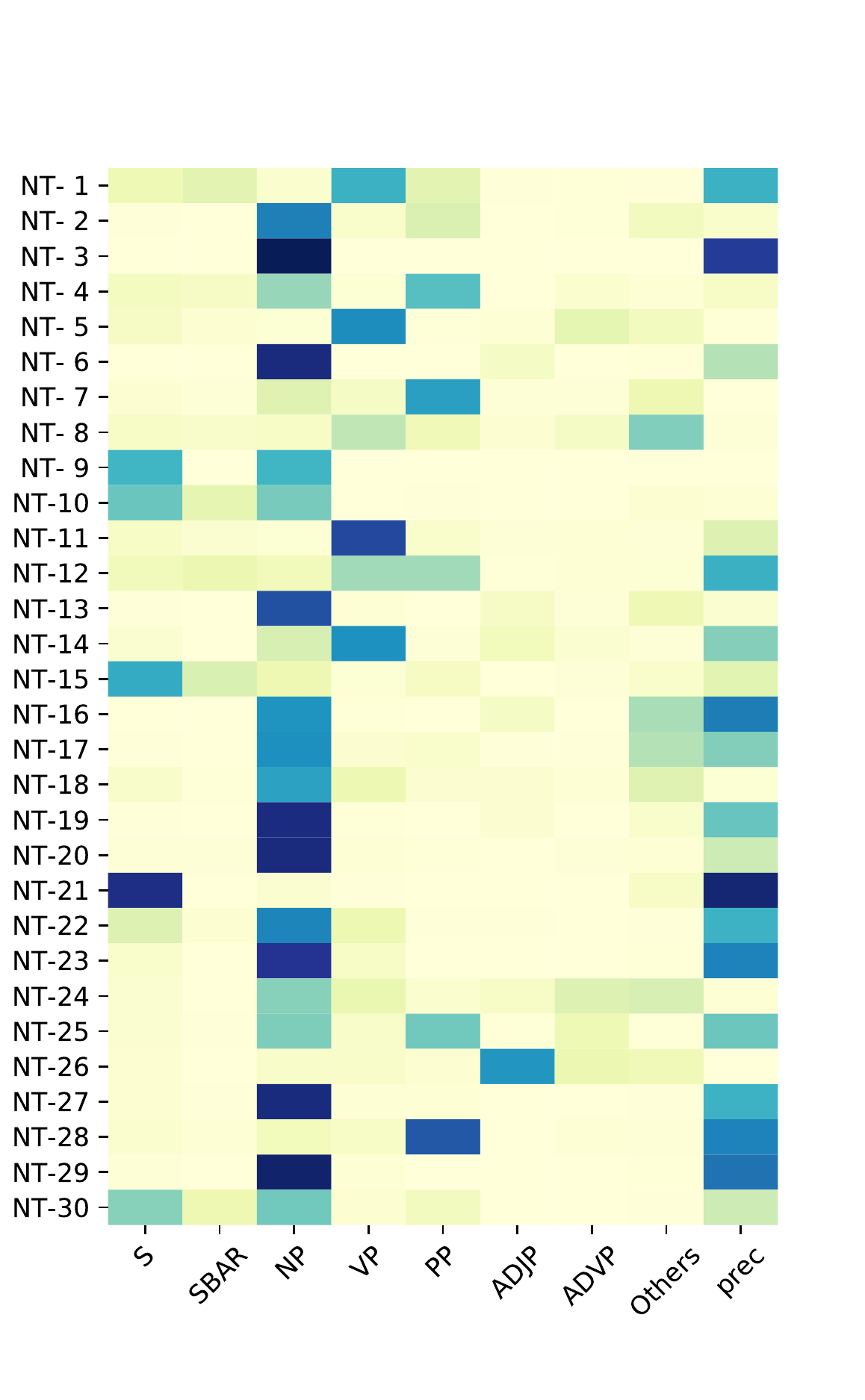}
    \caption{PCFG$_{\text{RoBERTa-large}}$}    
  \end{subfigure}

  \caption{ 
    Alignment of induced nonterminals of PCFG\subsetting{PLM} and PCFG\subsetting{Gold} on the entire PTB.
    The last column \texttt{prec} shows the precision that a nonterminal predicts a particular gold constituent.
  } 
  \label{fig: NT alignment}
\end{figure*}

\begin{figure*}[!htp]
  \centering
  \begin{subfigure}[b]{0.49\textwidth}
      \centering
      \includegraphics[width=\textwidth]{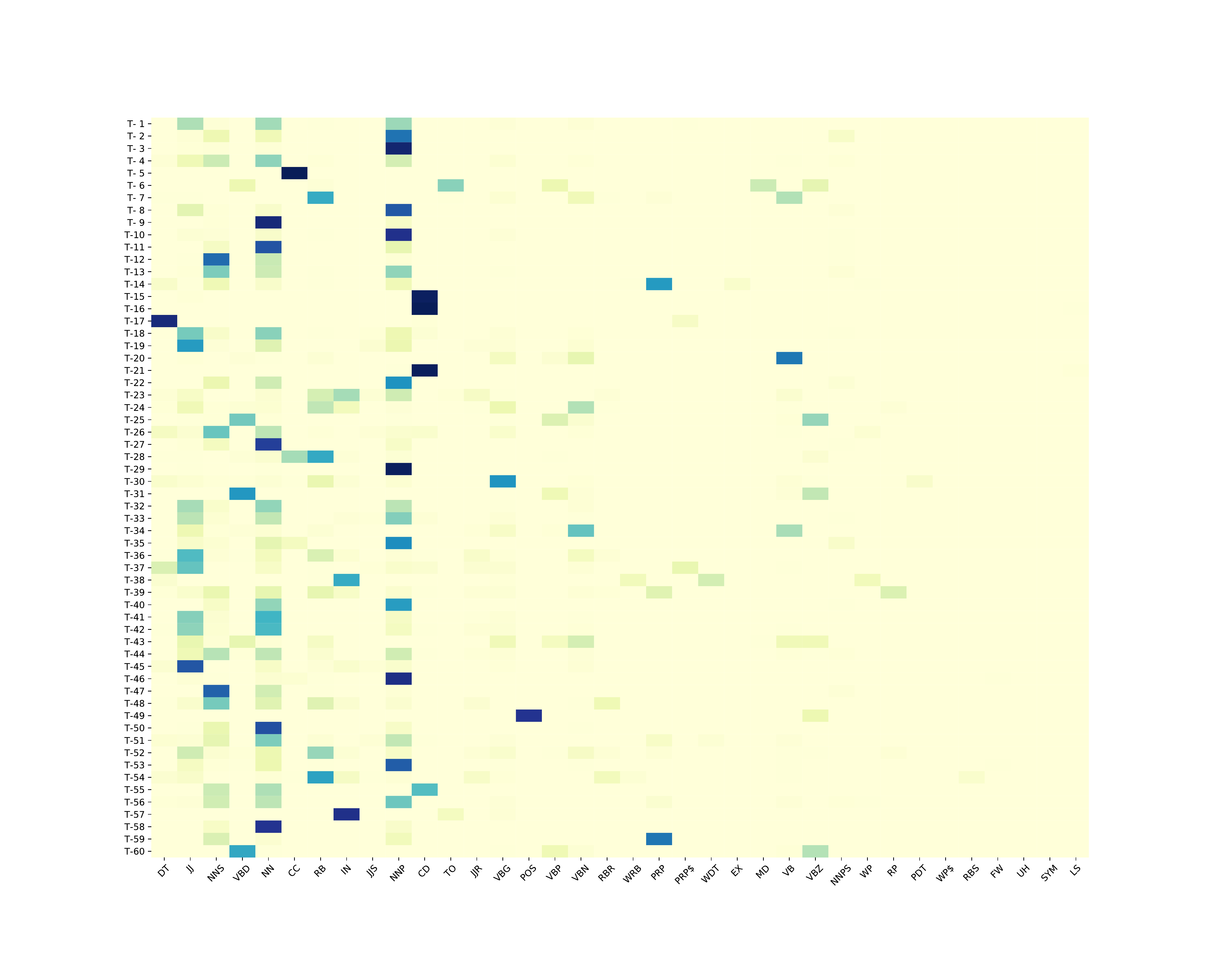}
      \vspace{-3em}
      \caption{PCFG$_{\text{Gold}}$}    
  \end{subfigure} \hfil
  \begin{subfigure}[b]{0.49\textwidth}
    \centering
    \includegraphics[width=\textwidth]{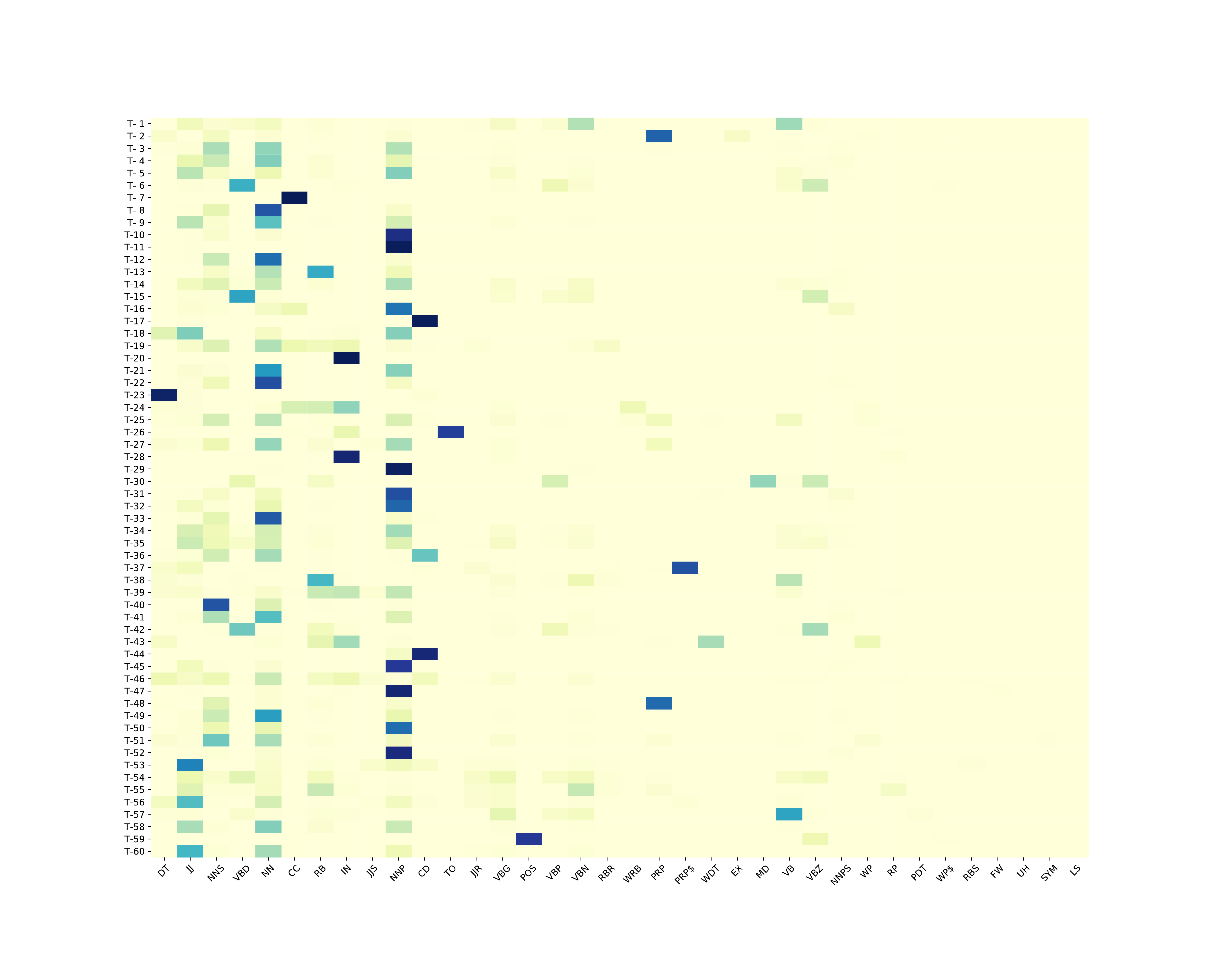}
    \vspace{-3em}
    \caption{PCFG$_{\text{BERT-base-cased}}$}    
  \end{subfigure} 

  \begin{subfigure}[b]{0.49\textwidth}   
      \centering 
      \includegraphics[width=\textwidth]{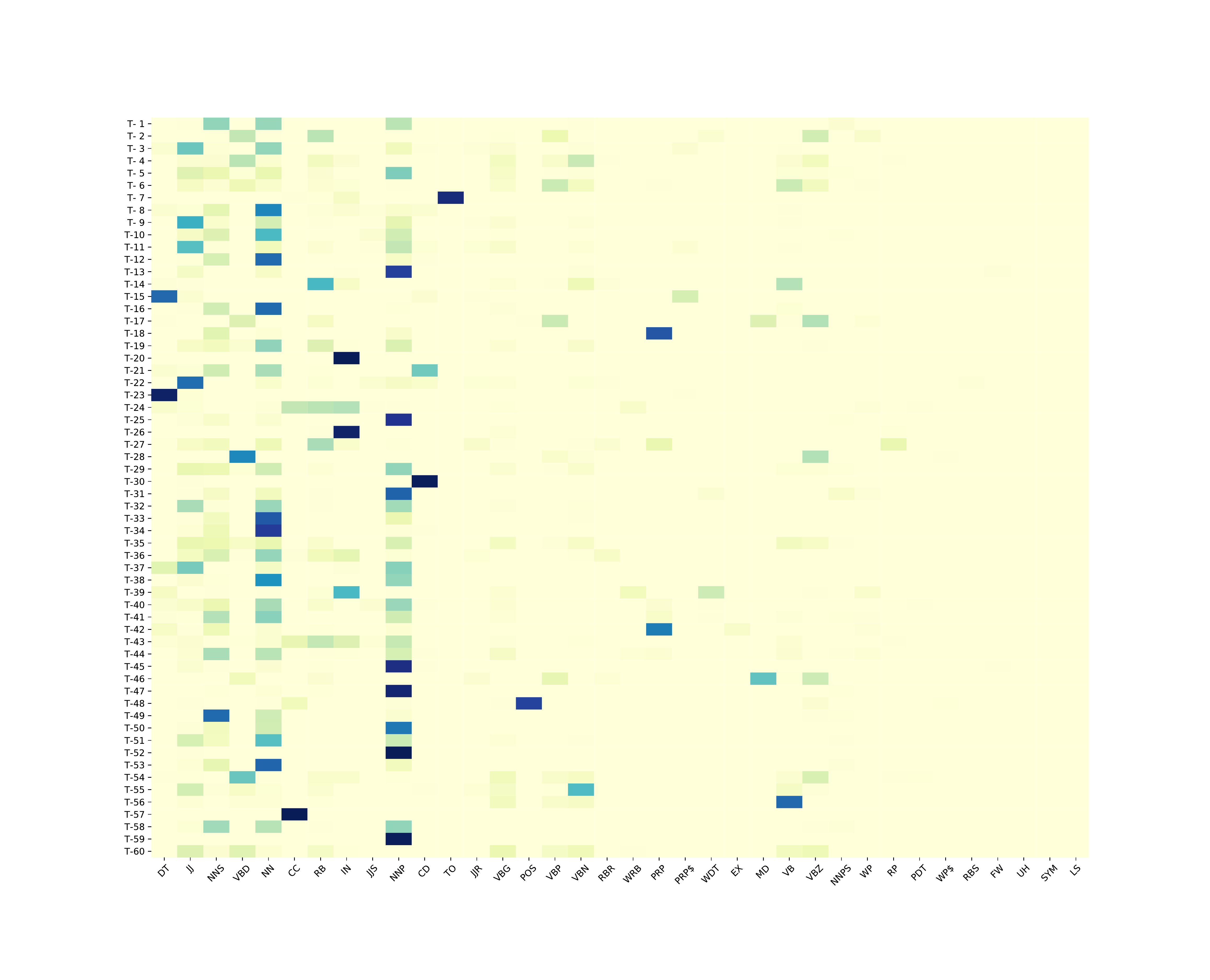}
      \vspace{-3em}
      \caption{PCFG$_{\text{BERT-large-cased}}$}    
  \end{subfigure} \hfil
  \begin{subfigure}[b]{0.49\textwidth}   
    \centering 
    \includegraphics[width=\textwidth]{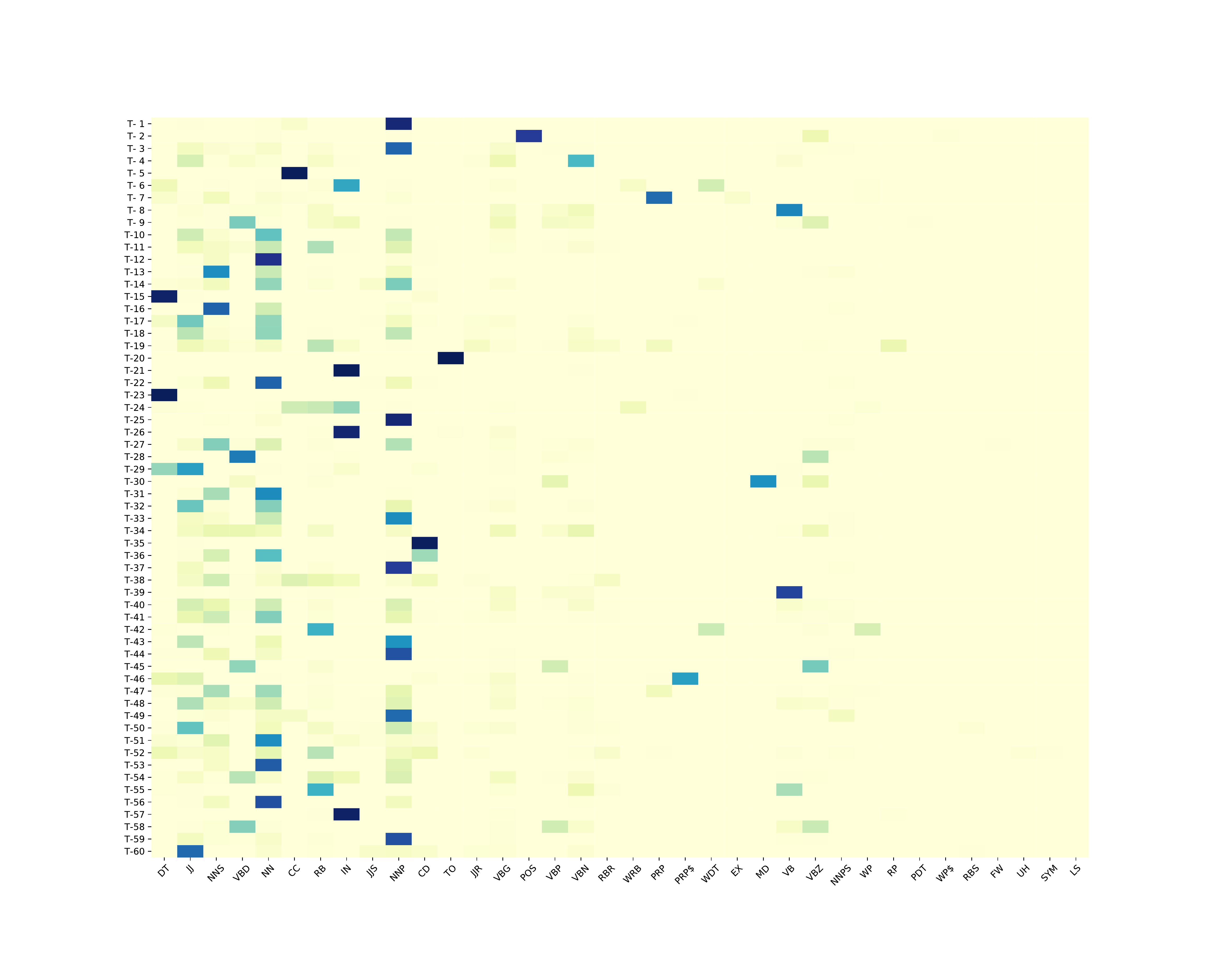}
    \vspace{-3em}
    \caption{PCFG$_{\text{XLNet-base-cased}}$}    
  \end{subfigure}

  \begin{subfigure}[b]{0.49\textwidth}   
    \centering 
    \includegraphics[width=\textwidth]{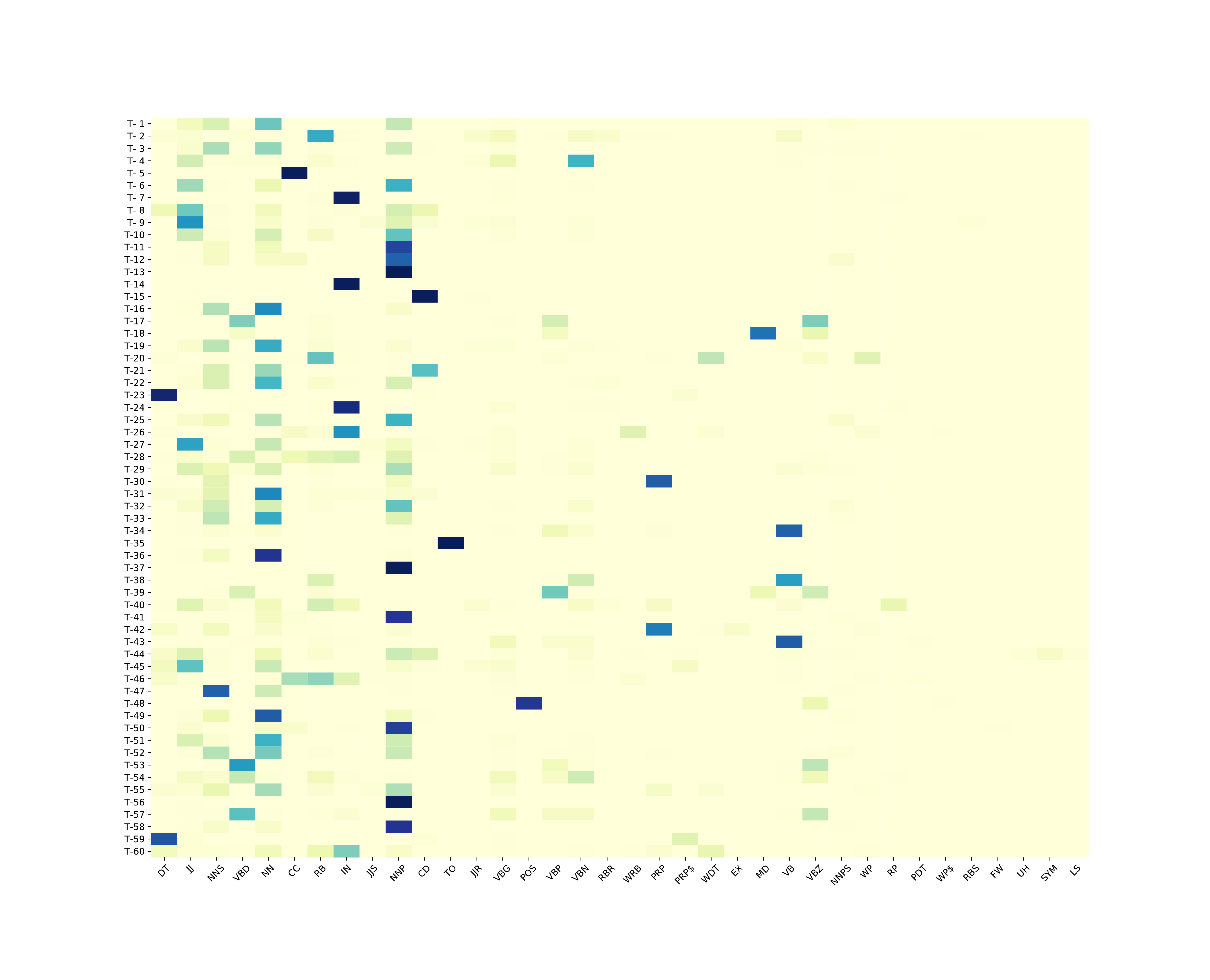}
    \vspace{-3em}
    \caption{PCFG$_{\text{XLNet-large-cased}}$}    
  \end{subfigure} \hfil
  \begin{subfigure}[b]{0.49\textwidth}   
    \centering 
    \includegraphics[width=\textwidth]{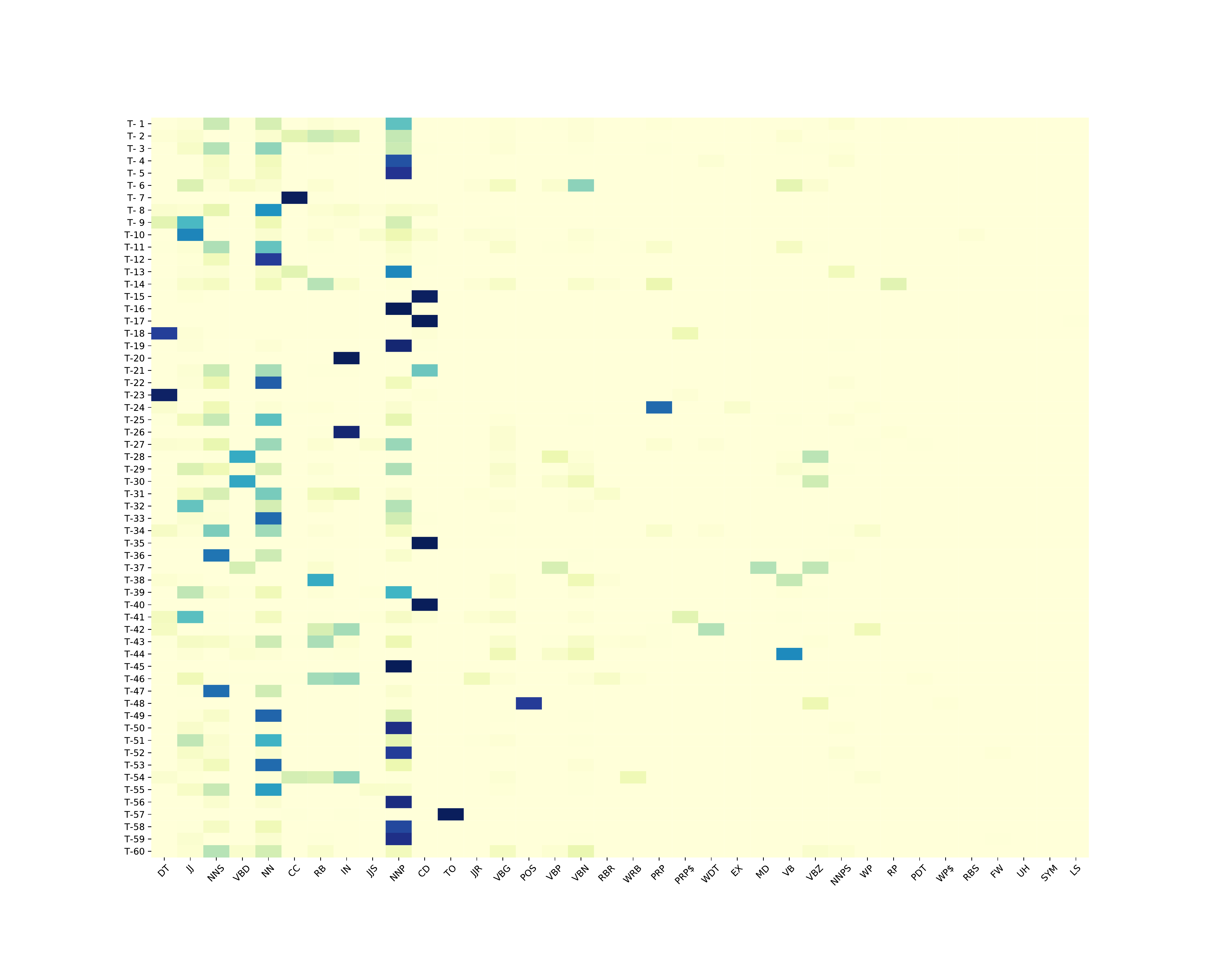}
    \vspace{-3em}
    \caption{PCFG$_{\text{RoBERTa-base}}$}    
  \end{subfigure} 
  
  \begin{subfigure}[b]{0.49\textwidth}   
    \centering 
    \includegraphics[width=\textwidth]{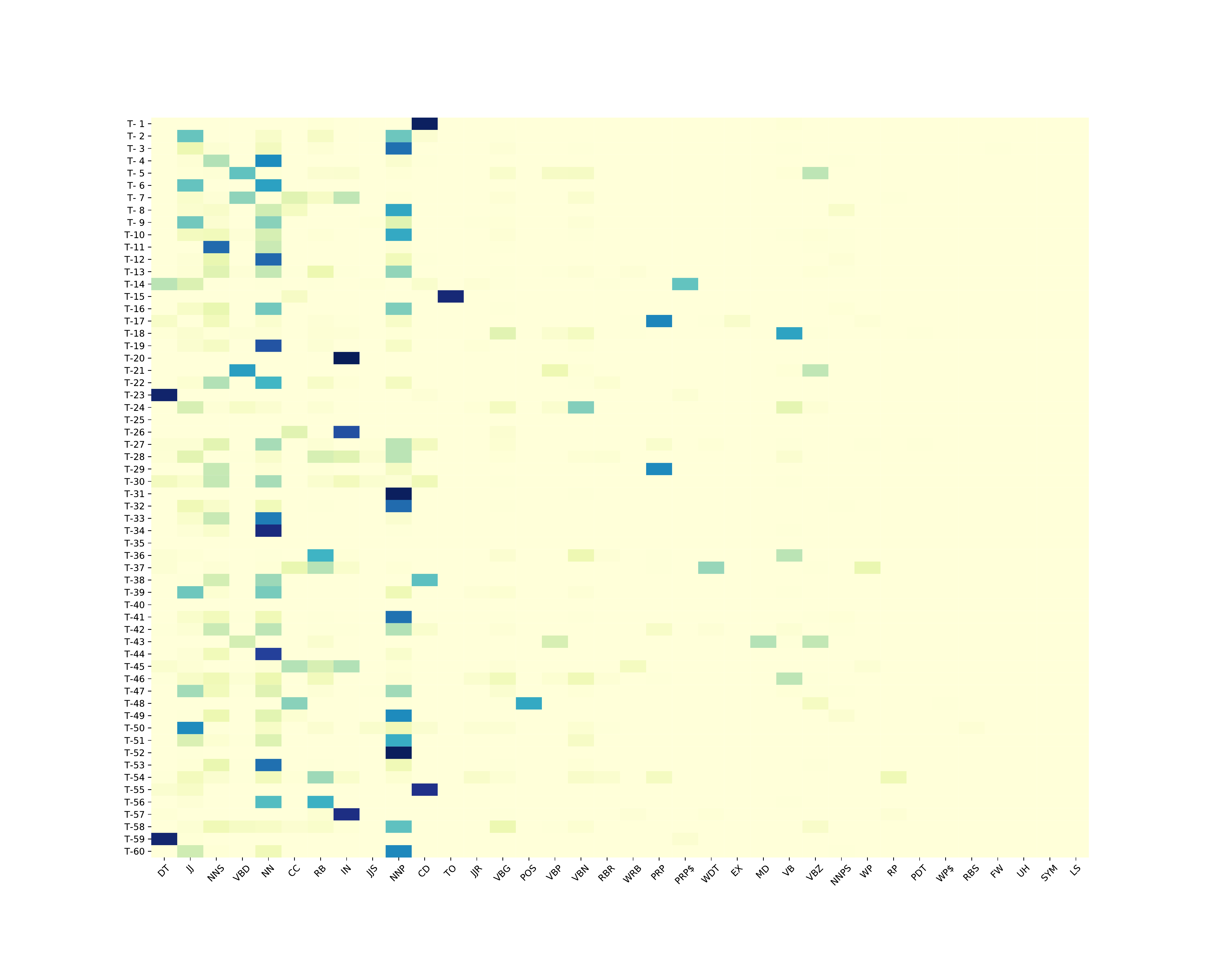}
    \vspace{-3em}
    \caption{PCFG$_{\text{RoBERTa-large}}$}    
  \end{subfigure}

  \caption{ 
    Alignment of induced preterminals (PoS tags) of PCFG\subsetting{PLM} and PCFG\subsetting{Gold} on the entire PTB.
  } 
  \label{fig: T alignment}
\end{figure*}

Since the recall scores in Table~\ref{tab: PTB tags} have shown ability of PLMs to identify different nonterminals, here we visualize the alignment between PCFG internal tags and corresponding gold labels in Figures~\ref{fig: NT alignment} and~\ref{fig: T alignment}. 
For the nonterminal alignment, some of the learned nonterminals clearly align to gold standard labels, in particular for frequent ones like NP and VP. 
Compared to PCFG$_\text{Gold}$ , PCFG$_\text{PLM}$ learns a more uncertain grammar and resulting in overall lower precision. 

But for the preterminal (PoS tag) alignment, no clear difference can be identified between PCFG$_\text{Gold}$ and PCFG$_\text{PLM}$.
This is consistent with the finding in Table~\ref{tab: POS and rule acc} that all PCFG$_\text{PLM}$ produce similar accuracies on preterminals as PCFG$_\text{Gold}$.

\subsection{Parse tree samples}
\label{appendix: tree samples}

\begin{figure*}[tb]
  \resizebox{1.0\linewidth}{!}{
  \begin{tabular}{c | c}
    \toprule
      \\
    \multicolumn{1}{l |}{\textbf{\LARGE Gold standard}}
    & \multicolumn{1}{l}{\textbf{\LARGE PCFG\subsetting{Gold}}}  \\
    \Tree [.{S} [.{NP-SBJ} [.{JJ} Foreign ] [.{NNS} bonds ] ] [.{VP} [.{VBD} surged ] [.{SBAR-TMP} [.{IN} as ] [.{S} [.{NP-SBJ} [.{DT} the ] [.{NN} dollar ] ] [.{VP} [.{VBD} weakened ] [.{PP-CLR} [.{IN} against ] [.{NP} [.{JJS} most ] [.{JJ} major ] [.{NNS} currencies ] ] ] ] ] ] ] ]
    &
    \Tree [.{NT-1 \redsub{S}} [.{NT-23 \redsub{NP}} [.{T-42 \redsub{JJ}} Foreign ] [.{T-47 \redsub{NNS}} bonds ] ] [.{NT-29 \redsub{VP}} [.{T-25 \redsub{VBD}} surged ] [.{NT-30 \redsub{PP}} [.{T-38 \redsub{IN}} as ] [.{NT-1 \redsub{S}} [.{NT-23 \redsub{NP}} [.{T-17 \redsub{DT}} the ] [.{T-27 \redsub{NN}} dollar ] ] [.{NT-29 \redsub{VP}} [.{T-25 \redsub{VBD}} weakened ] [.{NT-30 \redsub{PP}} [.{T-57 \redsub{IN}} against ] [.{NT-17 \redsub{NP}} [.{T-37 \redsub{JJ}} most ] [.{NT-10 \redsub{NP}} [.{T-41 \redsub{NN}} major ] [.{T-12 \redsub{NNS}} currencies ] ] ] ] ] ] ] ] ] \\ 
    \midrule
      \\
    \multicolumn{1}{l |}{\textbf{\LARGE PCFG\subsetting{BERT-base-cased}}} 
    &
    \multicolumn{1}{l}{\textbf{\LARGE PCFG\subsetting{BERT-large-cased}}} \\
    \Tree [.{NT-21 \redsub{S}} [.{NT-30 \redsub{NP}} [.{NT-4 \redsub{NP}} [.{NT-15 \redsub{S}} [.{NT-28 \redsub{NP}} [.{T-56 \redsub{JJ}} foreign ] [.{T-40 \redsub{NNS}} bonds ] ] [.{T-15 \redsub{VBD}} surged ] ] [.{T-26 \redsub{TO}} as ] ] [.{NT-9 \redsub{NP}} [.{NT-28 \redsub{NP}} [.{T-23 \redsub{DT}} the ] [.{T-22 \redsub{NN}} dollar ] ] [.{T-55 \redsub{RB}} weakened ] ] ] [.{NT-1 \redsub{VP}} [.{NT-5 \redsub{PP}} [.{T-28 \redsub{IN}} against ] [.{T-23 \redsub{DT}} most ] ] [.{NT-13 \redsub{NP}} [.{T-53 \redsub{JJ}} major ] [.{T-33 \redsub{NN}} currencies ] ] ] ]
    & 
    \Tree [.{NT-1 \redsub{S}} [.{NT-12 \redsub{NP}} [.{NT-21 \redsub{S}} [.{NT-15 \redsub{NP}} [.{T-3 \redsub{JJ}} foreign ] [.{T-49 \redsub{NNS}} bonds ] ] [.{T-4 \redsub{VBD}} surged ] ] [.{NT-25 \redsub{PP}} [.{T-26 \redsub{IN}} as ] [.{NT-15 \redsub{NP}} [.{T-23 \redsub{DT}} the ] [.{T-53 \redsub{NN}} dollar ] ] ] ] [.{NT-27 \redsub{VP}} [.{NT-26 \redsub{NP}} [.{T-4 \redsub{VBD}} weakened ] [.{T-26 \redsub{IN}} against ] ] [.{NT-14 \redsub{NP}} [.{T-11 \redsub{JJ}} most ] [.{NT-8 \redsub{NP}} [.{T-3 \redsub{JJ}} major ] [.{T-49 \redsub{NNS}} currencies ] ] ] ] ]
    \\
    \midrule
      \\
    \multicolumn{1}{l |}{\textbf{\LARGE PCFG\subsetting{XLNet-base-cased}}} 
    &
    \multicolumn{1}{l}{\textbf{\LARGE PCFG\subsetting{XLNet-large-cased}}} \\
    \Tree [.{NT-21 \redsub{S}} [.{NT-30 \redsub{NP}} [.{NT-3 \redsub{NP}} [.{T-17 \redsub{JJ}} foreign ] [.{T-16 \redsub{NNS}} bonds ] ] [.{T-58 \redsub{VBD}} surged ] ] [.{NT-1 \redsub{VP}} [.{NT-4 \redsub{S}} [.{NT-3 \redsub{NP}} [.{T-24 \redsub{IN}} as ] [.{NT-7 \redsub{NP}} [.{T-23 \redsub{DT}} the ] [.{T-22 \redsub{NN}} dollar ] ] ] [.{T-58 \redsub{VBD}} weakened ] ] [.{NT-12 \redsub{PP}} [.{T-26 \redsub{IN}} against ] [.{NT-14 \redsub{NP}} [.{T-50 \redsub{JJ}} most ] [.{NT-13 \redsub{NP}} [.{T-17 \redsub{JJ}} major ] [.{T-16 \redsub{NNS}} currencies ] ] ] ] ] ]
    & 
    \Tree [.{NT-1 \redsub{S}} [.{NT-12 \redsub{NP}} [.{NT-9 \redsub{NP}} [.{T-45 \redsub{JJ}} foreign ] [.{T-47 \redsub{NNS}} bonds ] ] [.{NT-11 \redsub{PP}} [.{NT-26 \redsub{NP}} [.{T-54 \redsub{VBD}} surged ] [.{T-7 \redsub{IN}} as ] ] [.{NT-28 \redsub{NP}} [.{T-23 \redsub{DT}} the ] [.{T-49 \redsub{NN}} dollar ] ] ] ] [.{NT-27 \redsub{VP}} [.{NT-26 \redsub{NP}} [.{T-54 \redsub{VBD}} weakened ] [.{T-7 \redsub{IN}} against ] ] [.{NT-4 \redsub{NP}} [.{T-23 \redsub{DT}} most ] [.{NT-23 \redsub{NP}} [.{T-9 \redsub{JJ}} major ] [.{T-16 \redsub{NN}} currencies ] ] ] ] ]
    \\
    \midrule
      \\
    \multicolumn{1}{l |}{\textbf{\LARGE PCFG\subsetting{RoBERTa-base}}} 
    &
    \multicolumn{1}{l}{\textbf{\LARGE PCFG\subsetting{RoBERTa-large}}} \\
    \Tree [.{NT-11 \redsub{S}} [.{NT-16 \redsub{NP}} [.{NT-9 \redsub{S}} [.{NT-19 \redsub{NP}} [.{T-51 \redsub{NN}} foreign ] [.{T-36 \redsub{NNS}} bonds ] ] [.{T-30 \redsub{VBD}} surged ] ] [.{NT-18 \redsub{PP}} [.{T-26 \redsub{IN}} as ] [.{NT-15 \redsub{NP}} [.{T-23 \redsub{DT}} the ] [.{T-22 \redsub{NN}} dollar ] ] ] ] [.{NT-27 \redsub{VP}} [.{NT-10 \redsub{VP}} [.{T-30 \redsub{VBD}} weakened ] [.{T-26 \redsub{IN}} against ] ] [.{NT-22 \redsub{NP}} [.{T-41 \redsub{JJ}} most ] [.{NT-19 \redsub{NP}} [.{T-41 \redsub{JJ}} major ] [.{T-47 \redsub{NNS}} currencies ] ] ] ] ]
    &
    \Tree [.{NT-21 \redsub{S}} [.{NT-30 \redsub{NP}} [.{NT-4 \redsub{PP}} [.{NT-15 \redsub{S}} [.{NT-13 \redsub{NP}} [.{T-9 \redsub{JJ}} foreign ] [.{T-22 \redsub{NN}} bonds ] ] [.{T-7 \redsub{VBD}} surged ] ] [.{T-57 \redsub{IN}} as ] ] [.{NT-23 \redsub{NP}} [.{T-23 \redsub{DT}} the ] [.{NT-13 \redsub{NP}} [.{T-39 \redsub{NN}} dollar ] [.{T-53 \redsub{NN}} weakened ] ] ] ] [.{NT-1 \redsub{VP}} [.{NT-7 \redsub{VP}} [.{T-26 \redsub{IN}} against ] [.{T-14 \redsub{PRP\$}} most ] ] [.{NT-18 \redsub{NP}} [.{T-50 \redsub{JJ}} major ] [.{T-4 \redsub{NN}} currencies ] ] ] ] \\
    \bottomrule
  \end{tabular}
  }
  \caption{
    Parse tree samples of gold standard, PCFG\subsetting{Gold}, and PCFG\subsetting{PLM}.
    The mapped tag (marked in red) for each anonymized nonterminal and preterminal is obtained via many-to-one mapping.}
  \label{fig: tree examples}
  \end{figure*}

In Figure~\ref{fig: tree examples}, we show parse trees obtained by PCFG\subsetting{Gold}, PCFG\subsetting{PLM} and the gold standard reference on a sample sentence.
In this sample, PCFG\subsetting{Gold} predicts the constituency tree structure accurately.
On the development set, PCFG\subsetting{Gold} reaches around 72 unlabeled $F_1$ score, as it is supervised by the PTB trees.
Although this is a low $F_1$-score, it is not untypical for PCFG-based models, which are limited by their insufficiently flexible rules and their lack of lexicalization. 
Also note that the oracle trees only yield 84.3 $F_1$.
PCFG\subsetting{PLM} perform worse than PCFG\subsetting{Gold} when compared against the gold tree. 
They are able to identify short NPs, but don't work well for larger constituents.
We also observe some frequent incorrect patterns which are also present in this example, e.g., grouping VBD with the preceding NP, or IN with the preceding VBD.

\end{document}